%% file: neurips_2026.tex
\documentclass{article}

\usepackage[preprint]{neurips_2026}

\usepackage[utf8]{inputenc} 
\usepackage[T1]{fontenc}    
\usepackage{hyperref}       
\usepackage{url}            
\usepackage{booktabs}       
\usepackage{amsfonts}       
\usepackage{pifont}         
\usepackage{nicefrac}       
\usepackage{microtype}      
\usepackage{xcolor}         

\usepackage{graphicx}       
\usepackage{amsmath, amssymb, amsthm, bm}   
\usepackage[most]{tcolorbox}    
\usepackage{multirow}       
\usepackage{tabularx}       
\usepackage[table]{xcolor}  
\usepackage{caption}        
\usepackage{subcaption}     
\usepackage{wrapfig}        
\usepackage{algorithm}      
\usepackage{algpseudocode}  

\newtcolorbox{promptbox}[1]{
    fontupper=\scriptsize,
    colback=gray!5!white,    
    colframe=gray!75!black,  
    fonttitle=\bfseries,     
    title={#1},              
    boxrule=0.5pt,           
    left=5pt, right=5pt, top=5pt, bottom=5pt
}

\newtheorem{theorem}{Theorem}[section]
\newtheorem{proposition}{Proposition}[section]

\newtheorem{lemma}[theorem]{Lemma}

\title{Democratizing Tool Learning with Environments \\ Fully Simulated by a Free 8B Language Model}

\author{
    Chenming Tang$^{\spadesuit\clubsuit}$
    \quad Hsiu-Yuan Huang$^{\spadesuit\clubsuit\diamondsuit}$\\
    \textbf{Weijie Liu}$^{\diamondsuit\dag}$\quad
    \textbf{Junqiang Zheng}$^\diamondsuit$\quad
    \textbf{Saiyong Yang}$^\diamondsuit$\quad \textbf{Yunfang Wu}$^{\spadesuit\clubsuit}$\thanks{Corresponding authors.} \\
    $^\spadesuit$National Key Laboratory for Multimedia Information Processing, Peking University \\ 
    $^\clubsuit$School of Computer Science, Peking University \quad
    $^\diamondsuit$LLM Department, Tencent \\
   \small{\href{jamydon.github.io}{\texttt{tangchenming@stu.pku.edu.cn}} \quad \href{mailto:jagerliu@tencent.com}{\texttt{jagerliu@tencent.com}} \quad \href{mailto:wuyf@pku.edu.cn}{\texttt{wuyf@pku.edu.cn}}}
}

\begin{document}

\setcounter{footnote}{1}

\maketitle

\setcounter{footnote}{1}
\renewcommand{\thefootnote}{\arabic{footnote}}

\begin{abstract}
  Reinforcement learning has become a prevalent paradigm for training tool calling agents, which typically requires online interactive environments. Existing approaches either rely on training data with ground truth annotations or require advanced proprietary language models (LMs) to synthesize environments that remain fixed once created. In this work, we propose TRUSTEE, a cost-friendly method for training tool calling agents with dynamic environments fully simulated by open-source LMs that can be as small as 8B, including task generation, user simulation, tool simulation, and trajectory evaluation, paired with an adaptive curriculum learning mechanism that controls task difficulty dynamically. Our empirical results show that TRUSTEE outperforms baselines which require extra external resources in most cases. These confirm that, with a sufficiently sophisticated design, even simulated environments with a local 8B LM as the backbone could set a strong baseline for tool learning. We hope our proposed paradigm could democratize tool learning and inspire future research on environment scaling with limited resources.
\end{abstract}

\section{Introduction}\label{sec:intro}

Tool calling, a.k.a. function calling, enables language models (LMs) to interact with the external environment and act as autonomous agents~\cite{toolqa, gorilla, react}. Earlier work teaching LMs to call tools (\textit{i.e.}, tool learning) typically adopts supervised fine-tuning (SFT)~\cite{toolformer, toolalpaca, toolllm} or single-turn reinforcement learning from verifiable rewards (RLVR)~\cite{toolrl,tool-n1,toolzero}, relying on annotated data. With the prevalence of agentic reinforcement learning (agentic RL)~\cite{search-r1}, recent research has been focusing on tool learning via multi-turn RL. However, due to the cost and limited availability of real-world users and tool infrastructures, there is a significant challenge in scaling the environment\footnote{In this paper, \textit{environment} refers to all the external components for the agent, including users, tools, verifiers, \textit{etc.}}, as this form of RL requires a large number of online interactions including calling many tools and dynamically interacting with the user, which also previously challenged evaluation~\cite{toolemu,stabletoolbench} and SFT data construction~\cite{agentscaler,button,toolace}.

One approach to environment scaling is to apply advanced LMs for environment simulation, but most of these works only partially simulate the environment (either the tools~\cite{randomworld,synthtools,from-word-to-world} or the users~\cite{userrl}) and rely on existing data or environments (\textit{i.e.}, supports some specific tasks only). For instance, Simia-RL~\cite{simia} simulates the responses of tools with LMs during interactions, but requires established data with annotated training tasks as the input (\textit{e.g.}, $\tau$-bench~\cite{tau-bench}) and does not support the multi-turn interaction between the agent and the user. Another approach is to make use of synthesized code and databases~\cite{envscaler,awm}, but such an approach requires a complex generation pipeline for high-quality synthesis using expensive proprietary LMs and the generated environments and tasks are fixed once created, which may be too easy for strong agents but overly difficult for weak ones.

To this end, we introduce TRUSTEE, \textbf{T}ool lea\textbf{R}ning \textbf{U}sing \textbf{S}imula\textbf{TE}d \textbf{E}nvironments, a cost-friendly method for training tool calling agents with a tool repository and an open-source LM as the only resources required. TRUSTEE uses an LM to generate the task, simulate the user, simulate tool responses, and evaluate the trajectories, which provides complete environments for tool learning while independent of any existing training data or tasks, human annotations or interactions, realistic executable tools, or verifiable environments engineered by experts or proprietary LMs. Moreover, we design an adaptive curriculum learning mechanism that takes advantage of the dynamically controllable nature of the simulated environments and tasks, which controls the difficulty of the tasks in various dimensions adaptively and ensures a smooth learning process for the tool calling agent.

Our experiments validate the effectiveness of TRUSTEE. For \textsc{Qwen3-8B-Base} and \textsc{Qwen3-4B}, TRUSTEE brings more positive gains than baselines in most cases. Moreover, the \textsc{Qwen3-8B} LM, when trained solely with environments simulated by itself, generally has better and more balanced performance than baselines which either require annotated data or need proprietary LMs to engineer the environments. This demonstrates that, with a sufficiently sophisticated design, simulated environments may be a powerful approach to tool learning that can even beat realistic data or synthesized verifiable environments. Analysis from different perspectives also helps us better understand the simulation quality, training stability, contributions of components, and cost efficiency.

Our contributions include:
\begin{itemize}
\item We introduce TRUSTEE, a paradigm of generating tasks and fully simulating environments with an LM for online tool learning.
\item We propose adaptive curriculum learning for tool learning, which dynamically keeps the tasks and environments at an appropriate difficulty for the policy. 
\item We empirically validate the effectiveness of TRUSTEE backboned by a free 8B LM, demonstrating tool learning does not necessarily require established data, executable tools, or verifiable environments from human experts or advanced proprietary LMs, moving a step towards democratized tool learning.
\end{itemize}

Our code is available at \url{https://github.com/JamyDon/TRUSTEE}.

\section{Method}

\subsection{Problem Formulation}
We formalize the tool calling agent task as a partially observable Markov decision process~\cite{survey}, defined by the tuple $\langle\mathcal{S}, \mathcal{A}, \mathcal{P}, \mathcal{R}, \gamma, \mathcal{O}\rangle$, where $\mathcal{S}$ is the state space, $\mathcal{A}=\mathcal{A}_\text{text}\cup\mathcal{A}_\text{tool}$ is the action space (textual response or tool calling), $\mathcal{P}$ is the state transition probability, $\mathcal{R}$ is the reward space, $\gamma$ is the discount factor, and $\mathcal{O}$ is the observation space. At the $t$-th POMDP step, the agent driven by an LM $\theta$ receives an observation $o_t$ based on the state $s_t\in\mathcal{S}$, takes an action $a_t\in\mathcal{A}$ based on $o_t$ and its policy $\pi_\theta$, receives a reward $r_t=\mathcal{R}(s_t,a_t)$, and the environment transitions per $\mathcal{P}(s_{t+1}\mid s_t,a_t)$. Instead of the standard discounted return, we solely use the final reward of each trajectory $\tau$ as the objective (see Appendix~\ref{app:adv} for details): $\mathcal{J}(\theta)=\mathop{\mathbb{E}}_{\tau\sim\pi_\theta}\left[r_\tau\right]$.

\subsection{Fully Simulated Environment}

\begin{figure}[htbp]
    \centering
    \includegraphics[width=0.9\linewidth]{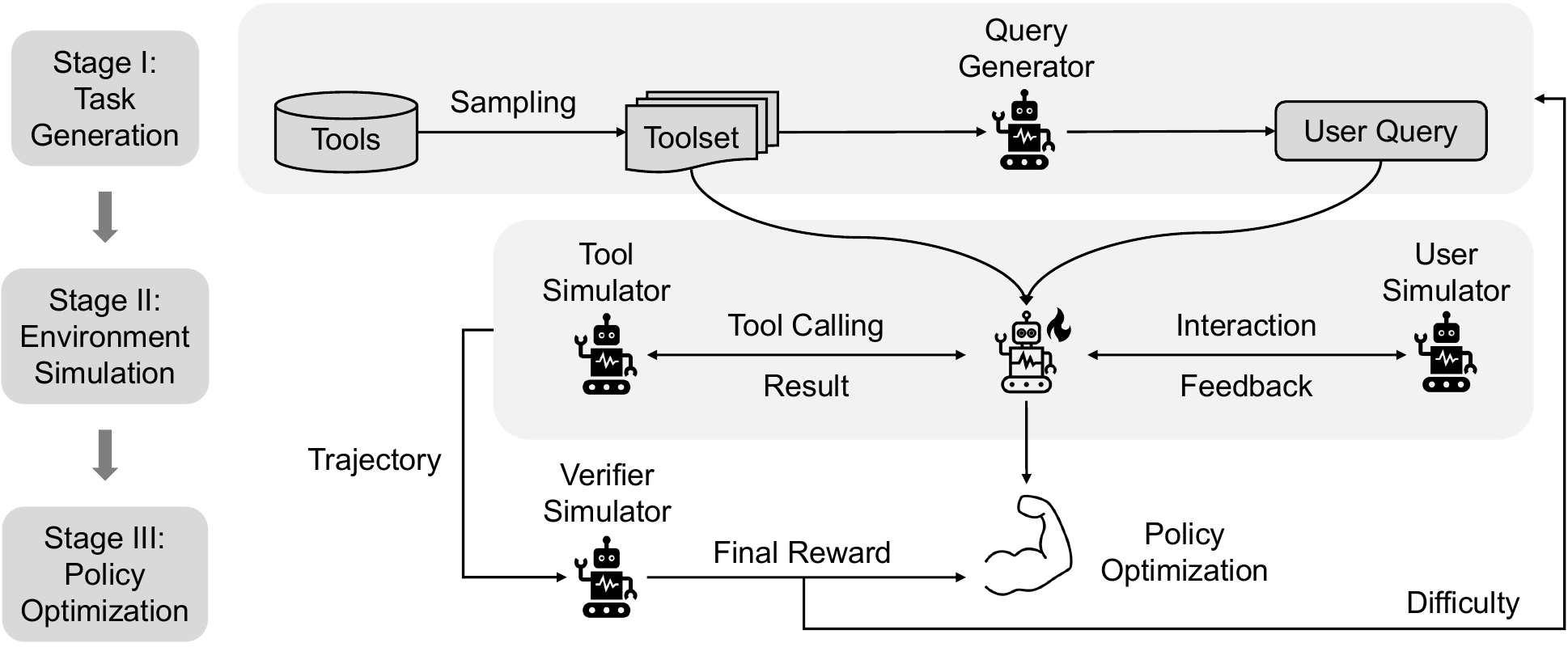}
    \caption{The overview of TRUSTEE.}
    \label{fig:method}
\end{figure}

Our fully simulated environment is demonstrated in Figure~\ref{fig:method}, which is backboned by an LM $\Phi$. There are three stages for each training iteration\footnote{In this section, we use the term ``iteration'' instead of ``step'' for training to distinguish from POMDP/interaction steps.}. The first generates tasks for this batch (Section~\ref{subsubsec:task-generation}). The second performs simulated interactions to collect trajectories (Section~\ref{subsubsec:simulated-interaction}). The last rewards the trajectories and optimizes the policy (Section~\ref{subsubsec:rewarding-optimization}). An algorithm description of TRUSTEE is in Appendix~\ref{app:algorithm}, and the prompt templates for the backbone LM can be found in Appendix~\ref{app:prompt}.

\subsubsection{Task Generation}\label{subsubsec:task-generation}
At the $n$-th training iteration, for each task $\mathcal{T}_i$, we first sample (denoted by $\zeta$) a subset $f$ of a large repository of tools $\mathcal{F}$, with the size of the subset determined by the task difficulty $\mathcal{D}_n$ (see Section~\ref{subsec:curriculum-learning}): $f=\zeta(\mathcal{F}, \mathcal{D}_n)$. In practice, the tool repository can be from various sources like existing datasets, crawled tools from the Internet, manually designed tool sets, \textit{etc.} Note that the tools only consist of basic textual information (names and parameters) and do not have to be actually executable functions.

Then, we perform task generation based on the backbone LM:
\begin{equation}
e_0\sim \mathcal{P}_\Phi(\cdot | \Omega_G),
\end{equation}
where $\Omega_G=\Omega(f, \mathcal{D}_n)$ is the structural prompt for task generation based on the tools $f$ and difficulty $\mathcal{D}_n$, and $e_0=(p, g, q, c)$ is the generated task including the user persona $p$, the user intent $g$ (\textit{i.e.}, the ultimate goal of the task), the first user query to trigger an interaction $q$, and the expected tools $c$ to be called to complete the task. Finally, after determining the agent's system prompt $h$ per $\mathcal{D}_n$ (see Section~\ref{subsubsec:task-difficulty}), we have the initial state $s_0$ based on all these information: $s_0=(f,p,g,q,c,h,\mathcal{D}_n)$.

\subsubsection{Simulated Interaction}\label{subsubsec:simulated-interaction}
We perform simulated interaction based on the generated tasks. The interaction environment includes a tool executor and a user, both simulated by the backbone LM $\Phi$.

For each task $\mathcal{T}_i$, we sample $M$ interaction trajectories. For each trajectory $\tau$, the policy agent $\pi_\theta$ begins the interaction after receiving the first observation $o_0 = \tau = (h, f, q)$, including the system prompt $h$, available tools $f$, and the first user query $q$. It interacts with both the tool executor and the user by generating actions based on the current trajectory $\tau$ (which is exactly the current observation):
\begin{equation}
a_t\sim\pi_\theta(\cdot|\tau),
\end{equation}
which will be appended to the trajectory: $\tau = \tau\oplus a_t$.

If $a_t$ is a tool call, the simulated tool will be called to generate a tool result based on the structural tool simulation prompt $\Omega_T=\Omega(f, g)$ which includes the tools $f$ and the user intent $g$. Otherwise, if $a_t$ is a textual response to the user, the simulated user will be called to generate a user message based on the structural user simulation prompt $\Omega_U=\Omega(f, g, p)$ which includes the available tools $f$, the user intent $g$, and the user persona $p$. The feedback is generated based on the corresponding prompt:
\begin{equation}
e_t\sim \mathcal{P}_\Phi(\cdot|\tau, \Omega_{T/U}),
\end{equation}
which will then be appended to the trajectory: $\tau = \tau\oplus e_t$.

The user simulator may choose to end the task by generating an empty response if it believes there is no need to perform more interactions. Otherwise, the interaction will continue until the sequence length or the interaction turn reaches a pre-defined limit.

Regarding the simulator, we do not require it to perfectly reproduce the real world. Reasonable and consistent interaction feedback is enough. For example, when learning to call a weather-search tool, whether the simulated API returns \texttt{Seattle, April 18: light rain, 52 °F} or the actual weather is different is less important than whether the agent learns to choose the right tool, pass the right city and date arguments, read the returned fields, and answer consistently. This view is also supported by ZeroSearch~\cite{zerosearch}, which trains search agents with simulated searches and reports that a 7B retrieval module achieves comparable performance to real search, while a 14B one can even surpass it. Therefore, in case the simulator is not an advanced LM (or even identical to the policy LM), it may be viewed as a limit on the training signal source, but not necessarily on the final policy.


\subsubsection{Rewarding and Optimization}\label{subsubsec:rewarding-optimization}

For each sampled trajectory $\tau$, when the interaction is finished, a verifier simulated by the backbone LM $\Phi$ is called to evaluate the whole trajectory and generates a final reward with the structural verifier simulation prompt $\Omega_V=\Omega(f, g, p, c, \mathcal{D}_n)$ based on the available tools $f$, the user intent $g$, the user persona $p$, the expected tools $c$, and the task difficulty $\mathcal{D}_n$: 
\begin{equation}
r\sim \mathcal{R}_\Phi(\cdot|\tau, \Omega_V).
\end{equation}
In this work, we have $r\in\{-1,0,1\}$ in our implementation. The trajectory receives a positive reward if all the criteria are met, a neutral reward if most criteria are met, and a negative reward if there are severe errors or the criteria are scarcely met.

We use GRPO~\cite{deepseekmath,search-r1} for optimization due to its stability and simplicity. Details are in Appendix~\ref{app:optimization}.


\subsection{Adaptive Curriculum Learning}\label{subsec:curriculum-learning}
An advantage of the fully simulated environment is that every component is dynamic and adjustable in an online manner, which naturally enables adaptive curriculum learning which keeps a moderate difficulty for the current policy and ensures good learnability.

\subsubsection{Task Difficulty}\label{subsubsec:task-difficulty}
The task difficulty involves the following dimensions. More details can be found in Appendix~\ref{app:task-difficulty}.

\paragraph{Number of Tools}
The number of tools sampled from $\mathcal{F}$ and the number of expected tool calls for task generation directly determine the overall difficulty. The more available tools and expected calls, the harder it is for the agent to make use of tools properly and finish the task precisely.

\paragraph{Number of Interaction Turns}
If a task requires more interaction turns, it is naturally more complex. Therefore, we use the number of expected turns, which will be passed to the LM $\Phi$ for task generation, to control the difficulty. Meanwhile, the maximum number of interaction turns also evolves dynamically with the expected turns to ensure efficient training.

\paragraph{System Prompt}
We expect the agent to learn to perform high-quality tool calling under a general system prompt and make the specificity of the system prompt gradually decrease during training. To this end, we design the system prompt $h$ at different degrees of specificity. The most specific one describes the overall multi-turn tool calling task in detail with hands-on instructions (a few paragraphs). Then, a more concise one presents the core notes (a few sentences) regarding the agentic task only. After that, a highly consolidated one states the task in a sentence (\textit{e.g.}, ``You are a helpful multi-turn dialogue assistant capable of leveraging tool calls to solve user tasks.''). Finally, the hardest system prompt is the most general one (\textit{e.g.}, ``You are a helpful assistant.''), with no special descriptions regarding any specific task. 

\paragraph{User}
Typically, a less expert user makes the task harder due to a lower level of knowledge~\cite{tau2-bench}, and an ambiguous user query usually confuses the agent and requires it to ask for further clarification. Therefore, we design three kinds of user personas (expert, beginner, and novice) together with three ambiguity degrees (clear, somewhat ambiguous, and highly ambiguous). Then, we use this as an input to the LM $\Phi$ for task generation which generates a more specific description of the user persona $p$ (which will be further passed to the user simulator) and a user query $q$ based on the requirements.

\paragraph{Rewarding Criteria}
Stricter evaluation criteria naturally make it harder for the agent to obtain a high reward. We increase the difficulty by gradually taking more criteria into consideration when calling the verifier $\Phi$, which gives a positive reward only when all the criteria are met. Concretely, we design five levels of criteria: meeting user intent, using expected tools correctly, no hallucination, no redundant trials, and efficiency. These draw a step-by-step path to a good tool calling agent.

\subsubsection{Difficulty Evolution}\label{difficulty-evo}
The difficulty is updated after each training iteration based on the overall reward of the finished batch. If the averaged reward $\bar{r}_n$ of training iteration $n$ is higher than a threshold $\eta_\text{high}$, the difficulty level for the next iteration $\mathcal{D}_{n+1}$ will increase by a step size $\delta$. Vice versa.
\begin{equation}
    \mathcal{D}_{n+1} = \begin{cases} 
\mathcal{D}_n + \delta & \text{if } \bar{r}_n > \eta_{\text{high}} \\
\mathcal{D}_n - \delta & \text{if } \bar{r}_n < \eta_{\text{low}} \\
\mathcal{D}_n & \text{otherwise}
\end{cases}.
\end{equation}
Then, all of the dimensions in Section~\ref{subsubsec:task-difficulty} will be derived based on $\mathcal{D}_{n+1}$. In this work, we design a large value range (\textit{e.g.}, from 1 to 100) for $\mathcal{D}$ and convert it to the scales of different dimensions. For example, since the range of user persona is 0-2 (0 for expert, 1 for beginner, and 2 for novice), the derived user persona level will be $\lfloor\mathcal{D}_{n+1}\times0.02\rfloor$. In this way, the user persona will be expert for difficulty from 1 to 49, beginner for difficulty from 50 to 99, and novice for difficulty 100. Since this mapping depends on the stepped design of each dimension, it is not treated as a hyper-parameter.

Since the difficulty works for every task in a batch, the batch may converge to a specific pattern of tasks, which hurts the diversity during training. For example, a detailed system prompt is always paired with an expert user. To this end, we employ a soft design of curriculum, which does not always follow the current difficulty strictly. To be specific, for each task, there is a probability of $\epsilon$ to sample the tool number, the expected tool callings, the interaction turns, the user persona, and the query ambiguity randomly within the current difficulty. In this way, the task patterns can be more diverse.

\subsection{Source of Improvement}\label{subsec:theory}
The essential engine of learning in TRUSTEE lies in the gap between the agent's limited observations and the environment's privileged knowledge (\textit{i.e.}, information asymmetry).

\begin{lemma}[Information Asymmetry]
The environment operates on a broader information set $\mathcal{I}_\text{env}=\{\tau, g, p, c, \mathcal{D}_n\}$ while the agent operates solely on $\mathcal{I}_\text{agent}=\{\tau\}$. Therefore, the environment may provide a training signal based on its privilege knowledge unknown to the agent.
\end{lemma}

Consequently, the improvement of TRUSTEE goes beyond a simple teacher-student distillation (from $\Phi$ to $\pi_\theta$) but can come from the structural design of the environment.

\begin{proposition}[Structural Improvement]\label{prop:structural-improvement}
Let $J(\pi) = \mathbb{E}_{\tau \sim \pi} [R(\tau)]$ denote the expected return of policy $\pi$. Given the environment backbone $\Phi$, the environment scaffolding $\mathcal{E}$ (including task generation, interaction, and rewarding), and the structural prompt set $\Omega$ for the backbone, the improvement $\Delta J=J(\pi_\theta)-J(\pi_0)$ can be decomposed into the transfer gain (from the backbone $\Phi$'s margin) and the structural gain (from the structural design including $\mathcal{E}$ and $\Omega$):
\begin{equation}
    \Delta J = \text{Gain}_{\text{trans}}(\Phi, \pi_\theta) + \text{Gain}_{\text{struct}}(\mathcal{E}, \Omega),
\end{equation}
where $\text{Gain}_{\text{struct}}>0$ even when $\Phi$ is not superior to $\theta$.
\end{proposition}

The detailed justifications for this proposition are provided in Appendix~\ref{app:theory}.

\section{Experimental Setup}\label{sec:setup}

\subsection{Data}

\paragraph{Evaluation Benchmarks}
We evaluate baselines and our method on the Berkeley Function Calling Leaderboard (BFCL) V4~\cite{bfcl} (we focus on two single-turn subsets ``Non-Live'' and ``Live'', and one multi-turn subset ``Multi-Turn'') and $\tau^2$-bench~\cite{tau2-bench}. The former tests general tool calling and the latter evaluates interactive tool calling in three realistic scenarios (``Airline'', ``Retail'', and ``Telecom''). Due to a limited budget, in our analytical experiments, we evaluate with BFCL V4 only.

\paragraph{Tool Repository}
We instantiate our tool repository $\mathcal{F}$ using the tools collected by ToolBench~\cite{toolllm}, which are representational state transfer (REST) APIs gathered from RapidAPI\footnote{https://rapidapi.com/hub} with detailed descriptions, spanning 49 diverse categories like social media, e-commerce, and weather. Details of data preprocessing are in Appendix~\ref{app:tool}. Experiments on stateful tools are in Appendix~\ref{app:stateful}.

\subsection{Language Models}
We focus on the \textsc{Qwen3}~\cite{qwen3} herd of language models, which is widely used for reinforcement learning in the community. To be specific, we experiment with \textsc{Qwen3-8B-Base}, \textsc{Qwen3-4B}, and \textsc{Qwen3-8B} as the agent LM $\theta$. Throughout our experiments, we use \textsc{Qwen3-8B} as the environment backbone LM $\Phi$ (see the discussion regarding the backbone LM in Appendix~\ref{app:teacher}).

\subsection{Baselines and Methods}
Our evaluated baselines and methods include:
(1) \textbf{Original Model}: the agent LM before training.
(2) \textbf{ToolRL}~\cite{toolrl}: a single-turn RLVR method using existing SFT data~\cite{xlam,toolace,hammer} with ground truth annotations.
(3) \textbf{Tool-N1}~\cite{tool-n1}: another single-turn RLVR method using existing SFT data~\cite{xlam,toolace}.
(4) \textbf{EnvScaler}~\cite{envscaler}: a method that generates environments via programmatic synthesis using proprietary LMs.
(5) \textbf{AWM}~\cite{awm}: another method that employs programmatic environment synthesis using proprietary LMs.
(6) \textbf{Simia-RL}~\cite{simia}: a method simulating the tools with proprietary LMs, requiring benchmark-specific environments.
(7) \textbf{TRUSTEE (ours)}: our method training the agent in environments simulated by an 8B open-source LM.

\subsection{Implementation Details}

We use a single node with 8$\times$ NVIDIA H20 GPUs. For TRUSTEE, 4 GPUs are used for agent training and the other 4 GPUs are used for deploying the simulation backbone LM; for other baselines, all the 8 GPUs are used for agent training. Since no extra hardware beyond the common budget is given to TRUSTEE, the cost is fair at the node level. Our training script is developed on the basis of \texttt{verl}~\cite{verl}. The simulation backbone LM is deployed using \texttt{vLLM}~\cite{vllm}. We extend the backbone LM's initial context size of 32,768 to 131,072 with a 4$\times$ YaRN~\cite{yarn} factor to avoid overlong trajectories.

For TRUSTEE, the step size of difficulty evolution $\delta$ is 3. The probability of soft curriculum $\epsilon$ is 0.5. The reward thresholds for difficulty update $\eta_\text{low}$ and $\eta_\text{high}$ are 0.0 and 0.5, respectively. The maximum numbers of tokens generated by the simulated user, tool executor, and verifier are all 2,048. The environment backbone LM operates with a frequency penalty of 0.1. The range of the number of tools is 1-10. The range of the expected tool callings is 1-3. The ranges of the expected turns, maximum user turns, maximum tool turns, maximum agent turns are 1-2, 1-3, 1-3, 1-6, respectively. To ensure the sampled tools from $\mathcal{F}$ are not overly discursive, we prioritize tools from the same category when sampling. We begin to sample from a new category only when all the tools have been selected in the previous one. To ensure stable tool learning, we include an extra format check before rewarding, in which the agent receives a negative reward if no tools are called throughout the interaction.

The details of evaluation can be found in Appendix~\ref{app:evaluation}. The details of baseline methods are presented in Appendix~\ref{app:baselines}. The hyper-parameters for training are listed in Appendix~\ref{app:hyper-parameters}.

\section{Results and Analysis}

\subsection{Main Results}

\input{tables/main}

The main end-to-end results are presented in Table~\ref{tab:main}.

For the pre-trained \textsc{Qwen3-8B-Base}, TRUSTEE brings significant improvements and secures the highest scores in 5 out of 6 subsets. It improves the base model by \textbf{10$\times$} and \textbf{15$\times$} on the two single-turn subsets of BFCL. On BFCL's multi-turn tool calling subset, it even improves the performance from \textbf{0.00} to \textbf{7.88} and outperforms the second highest baseline by \textbf{13$\times$}. On $\tau^2$-bench, it makes consistent improvements in all the subsets, with a significant \textbf{3.5$\times$} improvement on ``Telecom''. On the contrary, other baselines provide relatively limited gains, especially on the multi-turn subset of BFCL, where all the baselines fail to achieve meaningful progress.

The post-trained \textsc{Qwen3-4B} and \textsc{Qwen3-8B} already have a good capacity in tool calling, but TRUSTEE still brings generally positive gains in most cases. For \textsc{Qwen3-4B}, TRUSTEE performs generally better than all the baselines except EnvScaler, which requires a lot more time in training due to the complex environments (see Section~\ref{subsec:cost}). Nevertheless, for \textsc{Qwen3-4B}, no method provides consistent gains in all the cases, and we acknowledge TRUSTEE fails to improve on BFCL's Live and $\tau^2$-bench's Airline, indicating it is still challenging to further improve on an already post-trained LM.

For \textsc{Qwen3-8B}, TRUSTEE improves on all the subsets and secures the highest scores in 5 out of 6 cases, generally outperforming other baselines. Note that, for \textsc{Qwen3-8B}, the agent and the environment backbone share the same foundation LM. This demonstrates that with TRUSTEE, \textsc{Qwen3-8B} even benefits from itself when no annotated data or advanced LMs are involved, which empirically confirms the structural gain discussed in Section~\ref{subsec:theory}.

\subsection{Analysis of the Simulated Environment}\label{subsec:quality}

\input{tables/env_reliability}

Besides end-to-end performance, we also evaluate the overall reliability of TRUSTEE's simulated environment. We sample 100 trajectories via \textsc{Qwen3-8B} self-play in the simulated environment, and call \textsc{DeepSeek-V4-Pro}~\cite{DeepSeek-V4}, an advanced LM judge, to score (from 1 to 5) the simulation quality from various aspects. As shown in Table~\ref{tab:quality}, all the scores are above 3.5, suggesting a healthy simulation quality. The quality is especially good in task generation and user simulation, with scores over 4.5. As discussed in Section~\ref{subsubsec:simulated-interaction}, we do not need a perfect simulator, and the current quality is enough.

We also specifically evaluate the reliability of the verifier simulator. We use the test split of ToolRL's dataset \cite{toolrl}, in which each instance has a gold tool call and a rule-based scorer. We collect 410 trajectories whose correctness is mostly known by construction: gold positives, perturbations of gold calls inducing errors, and additional trajectories generated by \textsc{Qwen3-8B} labelled by the scorer. Then, we run \textsc{Qwen3-8B} as a reward verifier, and find it separates trajectories well. Against the rule-based scorer, the Spearman correlation is 0.70, accuracy is 0.87, F1 is 0.81 (precision is 0.72, recall is 0.93), and ROC-AUC is 0.89. The mean reward is 0.93 for correct and -0.08 for erroneous trajectories.

Finally, we perform manual audit on the only 20 trajectories generated by \textsc{Qwen3-8B} that were judged incorrect by the rule-based scorer, and find that half of them are actually correct. Some (seven) are due to the overly strict exact-match scorer. Others (three) are because the gold annotation is itself malformed. This shows that rule-based labels can be brittle in certain cases. Meanwhile, \textsc{Qwen3-8B} verifier has an accuracy of 0.85 on the 20 manually audited trajectories if we read a reward of +1 as good and 0 or -1 as bad. Therefore, the simulated verifier is generally reliable.

\subsection{Training Dynamics and Stability}

\begin{figure}[htbp]
    \centering
    \includegraphics[width=0.98\linewidth]{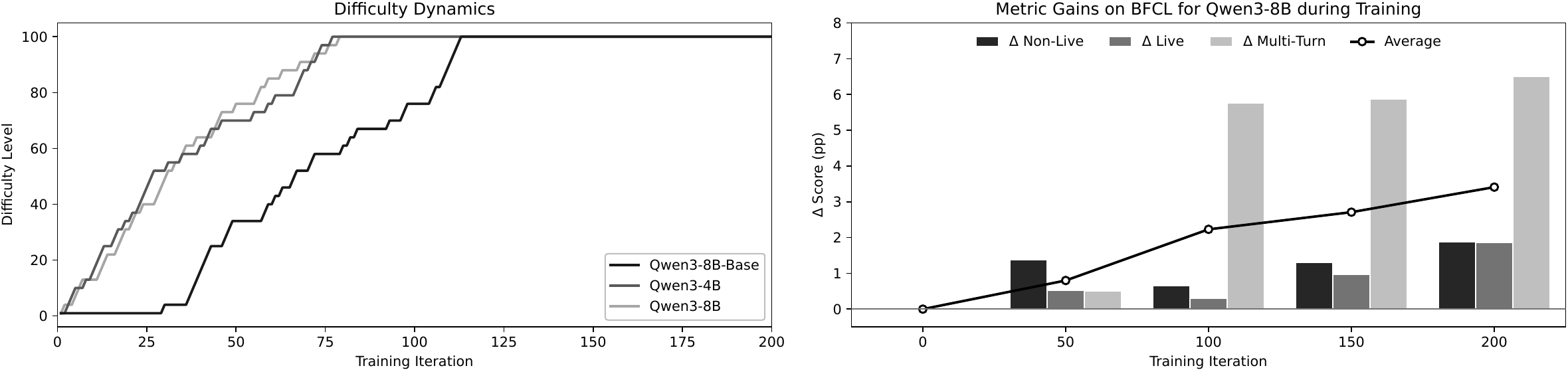}
    \caption{The dynamics of difficulty level (top) and metric gains (bottom).}
    \label{fig:dynamics}
\end{figure}

Figure~\ref{fig:dynamics} presents the dynamics of difficulty level and BFCL V4 metric gains. Throughout the training, the curriculum difficulty increases smoothly and reaches the upper bound around half of the training, and the overall performance is improving as the training proceeds. Meanwhile, according to our observation, the gradient norms remain bounded, the KL grows smoothly while remaining moderate, and the entropy decreases gradually. These suggest stable training throughout the process.

\subsection{Ablation Study}\label{subsec:ablation}

\input{tables/ablation_env}

We perform ablation studies on the environment's information set, adaptive curriculum learning, and hyper-parameters. The results on BFCL V4 with \textsc{Qwen3-8B} are in Table~\ref{tab:ablation-env}. These studies serve as stress tests of deliberately weakened variants, aiming to verify whether each component is necessary and understand the effect of hyper-parameters. It is therefore expected that the resulting method becomes less effective and can even hurt the original model, especially when the original \textsc{Qwen3-8B} model has already been post-trained. This supports the indispensability of these components and the importance of moderate hyper-parameters.

\paragraph{Ablation Study on the Environment's Information Set}
We analyze the three components of information used by the environment backbone LM: user intent $g$, user persona $p$, and expected tool calls $c$. We ablate these components by making them invisible to the backbone LM $\Phi$ for user, tool, and verifier simulation. As shown in Table~\ref{tab:ablation-env} (Part I), removing any one of the components hurts the performance on all the subsets. Without user intent, the environment backbone does not have sufficient information regarding the latent ground truth. Without user persona, the user simulator may not behave consistently and match the difficulty. Without expected calls, the verifier may deviate from the initial design of the task. These suggest the indispensability of all three components.




\paragraph{Ablation Study regarding Adaptive Curriculum Learning}
First, we ablate the whole adaptive curriculum learning module by fixing the difficulty at the highest level. Without a gradual learning process, the overall performance downgrades, especially on BFCL's Multi-Turn subset, where the scores decrease by \textbf{4.9} points for \textsc{Qwen3-8B}, shown by Table~\ref{tab:ablation-env} (Part II). Then, we ablate the soft curriculum design described in Section~\ref{difficulty-evo}, and the corresponding results confirm that a randomized difficulty prevents the agent LM from overfitting to undiversified tasks and environments and yields better performance. Finally, we perform an ablation study on the various dimensions for difficulty control: number of tools, number of interaction turns, system prompt, user, and rewarding criteria. We ablate each component by fixing it at the highest difficulty level throughout the training. As the results show, all five dimensions contribute positively on all three subsets of BFCL, which confirms all five dimensions are necessary to ensure a successful training outcome.






\paragraph{Effect of Hyper-parameters}
We analyze the effect of hyper-parameters in our method, including maximum number of available tools, maximum interaction turns, and the step size $\delta$ of difficulty evolution. We set three levels for each hyper-parameter: lower ($\downarrow$), default, and higher ($\uparrow$), with default as the value used in our main experiments. The detailed hyper-parameter values for each setting are presented in Appendix~\ref{app:setting-hparams}. As shown in Table~\ref{tab:ablation-env} (Part III), a moderate value for each hyper-parameter works best on all the subsets. Too many tools cause confusion, while too few tools cannot train the agent to handle complex tasks. Too many interaction turns cause unreliability for both the agent and the environment backbone due to limited long-horizon capability, yet too few turns make the environment too simple for the agent to learn to handle multi-turn interactions. Overly large step size $\delta$ makes the curriculum evolve too rapidly for the agent to establish a solid foundation in easier scenarios, and overly small $\delta$ fails to provide sufficient training for complex scenarios.

\subsection{Cost Analysis}\label{subsec:cost}
\input{tables/cost}

Table~\ref{tab:cost} presents the wall time consumed by training. TRUSTEE is within the same order of magnitude as ToolRL and Tool-N1 in training time and much shorter than EnvScaler under the same GPU budget. Note that the 4 GPUs for TRUSTEE's environment backbone are only active during the sampling phase. During optimization, they are idle, so there is no hidden extra budget for TRUSTEE; if anything, the agent itself is trained on fewer GPUs than the baselines that use all 8 GPUs for policy optimization.

Meanwhile, we only compare the training time here. If we count other costs, like data annotation or environment synthesis, other methods will be even more expensive (both in time and resource), while TRUSTEE needs none of these extra steps. Therefore, TRUSTEE is cost-efficient overall.

\section{Related Work}
\subsection{Tool Learning}
Tool calling, also known as function calling, allows LMs to obtain information from the external environment~\cite{toolqa,toolalpaca}, make use of helpful utilities~\cite{toolformer,gorilla}, and interact with the real world~\cite{react}, which converts generative assistants to intelligent agents.

Tool learning aims to enhance the tool calling capability of LMs via training. Earlier research focuses on SFT, which requires high-quality offline data. ToolAlpaca~\cite{toolalpaca} generates user queries and tool responses using LMs to synthesize single-turn data. BUTTON~\cite{button} and ToolACE~\cite{toolace} simulate both the user and tools to collect multi-turn trajectories. AgentScaler~\cite{agentscaler} instantiates tools as executable code and local databases and simulates the user with a user agent to collect multi-turn SFT data.

Search-R1~\cite{search-r1} first introduces agentic RL that allows LMs to learn to call tools (\textit{e.g.}, search engine APIs) via multi-turn reinforcement learning, which is superior in generalization. Many works perform RLVR with existing SFT data~\cite{toolrl,tool-n1,toolzero}. To enable truly multi-turn online learning, recent work constructs environments with LM-simulated tools~\cite{simia} or synthesized code and databases~\cite{autoforge,envscaler,awm}. Our work focuses on scaling the RL environment for tool learning with open-source LMs.

\subsection{Environment Scaling for LM Agents}
Real-world APIs are usually expensive and unstable~\cite{stabletoolbench} while customized tool calling infrastructures require a lot of expert labor, limiting the scalability of tool environments. Therefore, it is necessary to scale the environment for tool calling agents via simulation or synthesis. ZeroSearch~\cite{zerosearch} replaces the search engine API in Search-R1 with an LM. Simia-RL~\cite{simia} simulates tool responses in the online interactions with LMs. GTM~\cite{gtm} trains a specific LM to simulate tool responses. AutoForge~\cite{autoforge}, EnvScaler~\cite{envscaler} and AWM~\cite{awm} automatically generate user intents and build environments in the form of executable code and local databases for online reinforcement learning. Our work instantiates the whole environment as an LM and takes advantage of adaptive curriculum learning.

The novelty of our method lies in: (1) an end-to-end fully simulated environment for tool learning including task generation, user and tool simulation, and LM-based judgment; (2) an adaptive curriculum learning mechanism ensuring smooth training.

\section{Conclusion}
We have presented TRUSTEE, a cost-friendly method that simulates all the environment components for online tool learning. Empirical results show that TRUSTEE brings improvements in most cases and outperforms most of the compared baselines, which rely on either annotated data or environments synthesized by advanced proprietary LMs. Our work proves that, with a proper design, fully simulated environments can effectively train tool calling agents even with a local open-source 8B LM as the backbone while requiring no annotated data or tasks. Our work moves a step towards democratized tool learning and we hope it can inspire future research on environment scaling with limited resources.

Future work includes: (1) exploring other approaches to fine-grained credit assignment to take advantage of the turn-level rewards in our simulated environment and improve the training with denser supervision; (2) incorporating advanced entropy-controlling techniques to mitigate the entropy decrease during training; (3) incorporating a tool generation module that automatically generates various tools to free TRUSTEE from the reliance on a tool repository.

\section{Discussion}\label{sec:limitation-impact}

\paragraph{Limitations} While we provide a strong baseline for tool learning with an 8B LM, our work has certain limitations. First, the scalability of the simulated environment is still constrained due to the bounded capability of the simulation LM, which may not generate sufficiently complex tasks or sufficiently accurate rewards when the task involves a large number of tools and a lot of turns of interactions. Second, there are a lot of hyper-parameters introduced in our method (\textit{e.g.}, those in the adaptive curriculum learning), which makes it difficult to find an optimal setup, especially when the system is sensitive to hyper-parameters. Third, our experiments only validate TRUSTEE within the \textsc{Qwen3} family, and do not employ other model families like \textsc{Llama}~\cite{llama-3-2} due to infrastructure-side compatibility issues caused by different tool-call templates and parsers.

\paragraph{Broader Impact} This work democratizes tool learning with a methodology that simulates environments with open-source LMs. By reducing the reliance on human-annotated data, costly real-world infrastructures of executable environments, or expensive proprietary LMs, our work lowers the barrier for researchers to effectively and efficiently train tool calling agents. TRUSTEE can also be easily integrated into hybrid tool learning scenarios and may play the role of warm-up (\textit{e.g.}, using TRUSTEE for the initial training steps) or complement (\textit{e.g.}, using TRUSTEE for unavailable or costly tools). However, there may exist some malicious uses like modifying the prompts and tools to train agents with illegal tools (\textit{e.g.}, hacking, disinformation, harassment). We advocate for the intended uses of TRUSTEE and recommend deploying the trained agents in sandbox environments.




\begin{ack}
This work is partially supported by the Fundamental Research Funds for the Central Universities,
Peking University.
\end{ack}

\bibliographystyle{unsrt}
\bibliography{custom}

\clearpage


\appendix

\section{Algorithm of TRUSTEE}\label{app:algorithm}
We present the algorithm described in pseudo code of TRUSTEE in Algorithm~\ref{alg:trustee}. The soft curriculum mechanism is omitted for simplicity.

\input{algorithms/trustee}

\section{Prompt Templates}\label{app:prompt}

The prompt templates for task generation, tool simulation, user simulation, and verifier simulation are presented in Figure~\ref{fig:prompt-task-generation}, \ref{fig:prompt-tool-simulation}, \ref{fig:prompt-user-simulation}, and \ref{fig:prompt-verifier-simulation}, respectively.

\input{figures/prompt-task-generation}

\input{figures/prompt-tool-simulation}

\input{figures/prompt-user-simulation}

\input{figures/prompt-verifer-simulation}

\section{Theoretical Justifications}\label{app:theory}
Below we provide more detailed justifications for Proposition~\ref{prop:structural-improvement}. For simplicity, we assume $\Phi=\theta_0$, which means the environment backbone shares the same foundation model with the initial agent policy, to ablate the effect of standard distillation (\textit{i.e.}, $\text{Gain}_{\text{trans}}\approx0$). Then, we justify that $\Delta J>0$ by demonstrating the environment provides a training signal with higher information density than unguided autonomous self-play by the agent itself.

\paragraph{Distillation via Privileged Context.} The learning process can be formalized as the maximization of mutual information $I(\tau; z)$ between the agent's trajectory $\tau$ and the hidden knowledge $z = \{g, p, c, \mathcal{D}_n\}$ related to the ground truth. While the agent is restricted to $\mathcal{I}_\text{agent}$, the environment transition $\mathcal{P}_\Phi(e_t|\tau, \Omega_{T/U})$ is a function of the privileged information. Therefore, each observation $e_t$ acts as a signal in a noisy channel conveying knowledge regarding the ground truth. By optimizing the policy against the privileged verifier's reward conditioned on $z$, the agent minimizes the conditional entropy $H(z | \tau)$, distilling the privileged knowledge of the environment into a reactive policy. This ensures that the policy converges to the manifold of trajectories that are mutually informative of the ground truth goal even without direct access to $z$.

\paragraph{The Generator-Verifier Complexity Gap.} The computational asymmetry between generation and verification justifies the validity of the reward signal from the environment. Let $\pi_\theta(\tau)$ be the generative distribution and $V(\tau, z)$ be the verification function (the value of which is in $\{-1,0,1\}$ in our implementation). To solve the task, the agent has to search a massive action space to obtain a successful trajectory. On the contrary, the task for the verifier is deterministic: assessing whether the trajectory $\tau$ satisfies $z$. Consequently, the entropy of the environment's verification $H(V(\tau, z)|\mathcal{I}_\text{env})$ is strictly lower than that of the generative distribution of the agent $H(\pi_\theta(\tau)|\mathcal{I}_\text{agent})$ (\textit{i.e.}, $H_\text{ver} \ll H_\text{gen}$). Therefore, the verifier possesses a higher signal-to-noise ratio and acts as a filter truncating the agent's exploration distribution in the RL process. Even when $\Phi=\theta_0$, the lower complexity of the discriminative task ensures that the verifier provides a more informative gradient than an agent would provide through unguided self-play.

\paragraph{Manifold Expansion and Latent Unfolding.} Finally, we address the state-space coverage. Let $\mathcal{M}_\text{pre}$ be the manifold of knowledge acquired by the LM during pre-training. The zero-shot agent $\pi_\theta$ typically collapses to a high-probability sub-manifold $\mathcal{M}_\text{active} \subset \mathcal{M}_\text{pre}$. Therefore, the transition dynamics induced by the agent LM $\mathcal{P}_{\theta}$ is limited to high-probability scenarios. Nevertheless, the structural prompts $\Omega$ in the environment act as force-fields that shift the transition dynamics $\mathcal{P}_\Phi$, which expands the reachable state-space:
\begin{equation}
    \text{Vol}\left(\text{supp}(\mathcal{P}_\Phi | \Omega)\right) > \text{Vol}\left(\text{supp}(\mathcal{P}_{\theta})\right),
\end{equation}
where $\text{Vol}$ refers to the volume of the state-space manifold and $\text{supp}$ denotes the support (the set of reachable scenarios with reasonably large probability).
Due to this expansion, the agent is forced to interact with long-tail scenarios (\textit{e.g.}, a novice user or a highly ambiguous query), through which it unfolds latent capabilities that come from pre-training but are dormant in its default policy. Therefore, the environment can be seen to serve as a mechanism of alignment that projects the broad latent knowledge into a manifest policy.

In summary, the above three mechanisms facilitate a bootstrap effect collectively. By virtue of the simulated environment which is more informative and structurally diverse than the default reachable space of the agent, TRUSTEE enables an LM to transcend its initial performance limits. This theoretical framework demonstrates how an LM can act as its own supervisor, unfolding its latent potential into a specialized policy. The results of \textsc{Qwen3-8B} in our experiments also validate the structural gain empirically.

\section{Details of Optimization}\label{app:optimization}
We sample $M$ trajectories (each including $T_m$ interaction steps with $|a_t|$ tokens each) for each task $\mathcal{T}_i$ and apply multi-turn GRPO \cite{search-r1} with token-level loss \cite{dapo}:
\begin{equation}
\label{eq:grpo}
\mathcal{J}_{\text{GRPO}}(\theta)=
\mathop{\mathbb{E}}_{\mathcal{T}_i\sim\mathcal{T}}\left[\frac{1}{Z}\sum_{m=1}^{M}\sum_{t=0}^{T_m-1}\sum_{j=1}^{|a_t|}\left(\operatorname{CLIP}(r_{m,t,j}(\theta), \hat{A}_{m,t,j}, \varepsilon) -\beta\,\mathrm{D}_{\mathrm{KL}}(\pi_{\theta}\Vert\pi_{\text{ref}})\right)\right],
\end{equation}
where $Z$ is the normalization factor:
\begin{equation}\label{eq:normalization}
Z=\sum_{m=1}^{M}\sum_{t=0}^{T_m-1}|a_t|,
\end{equation}
$r_{m,t,j}(\theta)$ is the importance sampling ratio:
\begin{equation}\label{eq:ratio}
    r_{m,t,j}(\theta)
= \frac{\pi_{\theta}\bigl(x_{m,t,j}\mid \tau_{m, <t},\, x_{<j}\bigr)}
       {\pi_{\theta_{\mathrm{old}}}\bigl(x_{m,t,j}\mid \tau_{m, <t},\,x_{<j}\bigr)},
\end{equation}
\(\hat{A}_{m,t,j}\) is the group-relative advantage based on the final reward only:
\begin{equation}\label{eq:adv}
\hat{A}_{m,t,j}
= \frac{r_m \;-\;\mathrm{Mean}\bigl(\{r_m\}_{m=1}^M\bigr)}
       {\mathrm{Std}\bigl(\{r_m\}_{m=1}^M\bigr)},
\end{equation}
and $\mathrm{D}_{\mathrm{KL}}(\pi_{\theta}\Vert\pi_{\text{ref}})$ is the KL penalty term.

\section{More Implementation Details}\label{app:implementation-detail}
\subsection{Details of Task Difficulty}\label{app:task-difficulty}
\input{tables/difficulty}

The detailed descriptions of the different difficulty dimensions at all levels are presented in Table~\ref{tab:difficulty}. Note that the rewarding criteria are applied in an incremental manner. The criteria at level $l$ should include all those from level $0$ to level $l$.

\input{figures/system-prmopt-detailed}

\input{figures/system-prompt-concise}

\subsection{Details of Tool Repository}\label{app:tool}

\input{figures/prompt-tool-filter}

We preprocess the raw tool repository with \textsc{Qwen3-Next-80B-A3B-Instruct}\footnote{https://huggingface.co/Qwen/Qwen3-Next-80B-A3B-Instruct}, discarding low-quality tools. The prompt template for the LM is presented in Figure~\ref{fig:prompt-tool-filter}.

\subsection{Details of Evaluation}\label{app:evaluation}
For BFCL V4, we follow the default evaluation setup and do not further extend the maximum context size of the LM for long-context scenarios. Please note that our evaluated BFCL V4 scores may not be directly comparable to the official leaderboard\footnote{\url{https://gorilla.cs.berkeley.edu/leaderboard.html}}, because we are using a newer version of the evolving BFCL V4 (committed in March 2026) and inference backends may differ (we are using local \texttt{vLLM} while the leaderboard does not specify this).

For $\tau^2$-bench, we adopt \textsc{GPT-4.1} as the user simulator following \cite{tau2-bench}. The maximum number of interaction turns is 64 in our experiments due to limited budget. The cost for each run is about \$30.

\subsection{Details of Baselines}\label{app:baselines}
\paragraph{ToolRL}
We reproduce ToolRL based on their released scripts\footnote{\url{https://github.com/qiancheng0/ToolRL}} (we do not directly run theirs due to model incompatibility) with aligned maximum sequence length. We ran the training for 500 steps with a batch size of 32 to align the overall training time. However, the training for \textsc{Qwen3-8B} collapsed soon after 400 steps. Therefore, we evaluate the checkpoint of 400 steps for \textsc{Qwen3-8B}.

\paragraph{Tool-N1}
We reproduce Tool-N1 based on their released scripts\footnote{\url{https://github.com/NVlabs/Tool-N1}} (we do not directly run theirs due to model incompatibility) with aligned maximum sequence length. We ran the training for 500 steps with a batch size of 32 to align the overall training time.

\paragraph{AWM}
We evaluate their released checkpoint\footnote{\url{https://huggingface.co/Snowflake/Arctic-AWM-8B}} trained using \textsc{Qwen3-4B} and \textsc{Qwen3-8B} directly because no training script is released and we cannot reproduce the method. There is no available checkpoint for \textsc{Qwen3-8B-Base}, thus there is no result of AWM for this LM.

\paragraph{EnvScaler}
We train the LM with the provided scripts\footnote{\url{https://github.com/RUC-NLPIR/EnvScaler}} and the default hyper-parameters. We focus on the pure RL setting and thus train the model from scratch without SFT, which is in line with TRUSTEE and other baselines.

\paragraph{Simia-RL}
We evaluate their released checkpoint\footnote{\url{https://huggingface.co/Simia-Agent/Simia-Tau-SFT-RL-Qwen3-8B}} trained using \textsc{Qwen3-8B} on $\tau$-bench directly because re-running the training was infeasible within budget due to Simia-RL's reliance on expensive proprietary LMs. Meanwhile, there is no available checkpoint for \textsc{Qwen3-8B-Base} or \textsc{Qwen3-4B}, thus there is no result of Simia-RL for these LMs.

\subsection{Hyper-parameters}\label{app:hyper-parameters}
\input{tables/hyper-parameters}
The hyper-parameters for training are listed in Table~\ref{tab:hyper-parameters}. The batch size is only 16 because a larger batch size requires a higher level of concurrency which may cause the local \texttt{vLLM} server for the simulation LM to be overloaded.

\subsection{Details for Hyper-parameter Ablations}\label{app:setting-hparams}
\input{tables/setting_hparams}

The detailed values of different hyper-parameters used in Section~\ref{subsec:ablation} are presented in Table~\ref{tab:setting-hparam}.

\section{Extended Experimental Results}
\subsection{Confidence Intervals of the Main Results}\label{app:ci}

\input{tables/ci}

We report the main results with 95\% confidence intervals in Table~\ref{tab:ci}. For \textsc{Qwen3-8B-Base} on BFCL, TRUSTEE's intervals do not overlap with the original model's, supporting its significant gain. Yet there are certain overlaps of intervals when the policy LM has already been post-trained (\textsc{Qwen3-4B} and \textsc{Qwen3-8B}), for both TRUSTEE and other compared methods. From a relative perspective, TRUSTEE remains the strongest or tied strongest method by point estimate in 12 out of 18 model-benchmark settings in the table, while using no annotated data and no proprietary LM.

\subsection{Effect of Turn-level Fine-grained Reward}\label{app:adv}

\input{tables/adv}

We explore integrating the turn-level fine-grained reward given by the backbone LM simulating the user or tool executor. The advantage for the $t$-th turn is given by:
\begin{equation}
    \hat{A}_{t} = \begin{cases} 
\frac{r_t+1}{2} & \text{if } r_t\ge0 \\
\frac{r_t-1}{2} & \text{if } r_t<0
\end{cases},
\end{equation}
where $r_t$ is the reward of the $t$-th turn. Therefore, a positive turn-level reward leads to a positive advantage, while a negative reward leads to a negative advantage. Meanwhile, a reward with a higher absolute value leads to a larger-magnitude advantage.

We replace the default sequence-level advantage with this turn-level one and experiment with \textsc{Qwen3-4B}. As shown in Table~\ref{tab:adv}, the fine-grained advantage does not necessarily provide further improvements and even exhibits a degradation on the Multi-Turn subset. This indicates that the default sequence-level advantage already provides reliable performance. Therefore, we adopt the sequence-level advantage by default in our experiments.

Nevertheless, this is a na\"ive design of turn-level advantages, which does not guarantee the zero-mean property and may not support high-quality training. We leave a more advanced turn-level advantage calculation for future work.

\subsection{Effect of the Environment Backbone LM}\label{app:teacher}

\input{tables/teacher}

We replace the environment backbone with \textsc{Qwen3.5-397B-A17B-FP8}\footnote{https://huggingface.co/Qwen/Qwen3.5-397B-A17B-FP8}, which is a more advanced LM (but quantized), to analyze the effect of the backbone LM. Due to the large size of this LM, we use 8 GPUs for serving the environment backbone and another 8 GPUs for training the agent.

As shown in Table~\ref{tab:teacher}, the effect of the backbone LM is complex. The more advanced backbone LM exhibits improvements in the Live subset for \textsc{Qwen3-8B-Base} and in the Multi-Turn subset for \textsc{Qwen3-4B}, but fails to bring further gains in other cases. This may be due to the downgraded performance of the FP8-quantized LM and certain limitations in sparse models. Anyway, we can still conclude that \textsc{Qwen3-8B} is already a good choice for the backbone of TRUSTEE, which brings improvements with high efficiency.

Due to a limited budget, we cannot afford to further experiment with other advanced backbones, especially proprietary LMs. In our experiments, each complete training requires an order of magnitude of \(10^{8}\) tokens from the backbone LM approximately, which demands an extremely high cost.

\subsection{Simulating Stateful Tools}\label{app:stateful}

Like the simulation quality evaluation in Section~\ref{subsec:quality}, we use the stateful tools from EnvScaler~\cite{envscaler} as the tool repository and sample 100 complete trajectories via \textsc{Qwen3-8B} self-play in the simulated environment for \textsc{DeepSeek-V4-Pro} to judge the simulation quality across all aspects (from 1 to 5). The overall simulation quality is also healthy, with all aspects scoring above 4.0. To be specific, the scores for task generation, user simulation, tool simulation, final reward, internal consistency, and overall fidelity are 4.65, 4.37, 4.82, 4.20, 4.72, and 4.36, respectively.

\input{tables/stateful}

To further study the simulation performance of stateful tools, we also train \textsc{Qwen3-8B-Base} with the stateful tools used above under our main experimental settings and report the end-to-end results on BFCL in Table~\ref{tab:stateful}. The stateful setting still brings significant improvement to the base model. Nevertheless, its gains are smaller than the default stateless setting, especially on the multi-turn subset, which may reflect both the harder stateful-tool setup and the narrower tool and task coverage due to a much smaller tool repository.






\section{Computational Resources}\label{app:compute}
All our experiments are conducted on a machine with TencentOS Server 4, 384 AMD$^\circledR$ EPYC\texttrademark{} 9K84 96-Core Processor CPUs and 2.2TiB memory. We use 8$\times$ NVIDIA H20 GPUs for both training and evaluation for all the experiments (except for the experiments in Appendix~\ref{app:teacher}).

\section{Use of AI Assistants}\label{app:llm}
The algorithmic design and core methodology of this work were derived through manual research and human reasoning. We use AI assistants solely for wording, editing, formatting, and coding purposes. They do not impact the core methodology, scientific rigor, or originality of the research.

\section{Use of Scientific Artifacts}\label{app:license}
We cite all the creators of scientific artifacts we use in this paper. Licenses and URLs of these scientific artifacts are shown in Table~\ref{tab:license}. Our use of these is consistent with their intended use.

\input{tables/license}


\newpage
\input{checklist.tex}

\end{document}

%% file: tables/main.tex
\begin{table}[htbp]
\small
  \centering
  \caption{Main results. The highest results for each LM are \textbf{bolded}. ``DF'' refers to data-free (\textit{i.e.}, no established data) while ``OP'' indicates using open-source LMs only. The BFCL results of methods with ``$\dag$'' are evaluated under the direct prompting setting instead of the standard native function calling (FC) setting because they score nearly 0 on all the categories under the default FC setting.}
    \begin{tabular}{lcccccccc}
    \toprule
    \multicolumn{1}{c}{\multirow{2}[2]{*}{\textbf{Method}}} & \multirow{2}[2]{*}{\textbf{DF}} & \multirow{2}[2]{*}{\textbf{OP}} & \multicolumn{3}{c}{\textbf{BFCL V4}} & \multicolumn{3}{c}{\textbf{$\tau^2$-bench}} \\
\cmidrule(lr){4-6}\cmidrule(lr){7-9}          &       &       & \textbf{Non-Live} & \textbf{Live} & \textbf{Multi-Turn} & \textbf{Airline} & \textbf{Retail} & \textbf{Telecom} \\
    \midrule
    \textsc{Qwen3-8B-Base} & -     & -     & 8.38  & 4.74  & 0.00  & 26.00 & 4.39  & 1.75 \\
    \;ToolRL & \ding{55}     & \checkmark     & 40.83 & 56.33 & 0.25  & 30.00 & \textbf{6.14} & 5.26 \\
    \;Tool-N1$^\dag$ & \ding{55}     & \checkmark     & 37.85 & 29.53 & 0.12  & \textbf{32.00} & 4.39  & 4.39 \\
    \;EnvScaler$^\dag$ & \checkmark     & \ding{55}     & 75.29 & 62.03 & 0.62  & 28.00 & \textbf{6.14} & 0.00 \\
    \rowcolor[rgb]{ .851,  .851,  .851} \;TRUSTEE & \checkmark     & \checkmark     & \textbf{83.92} & \textbf{72.54} & \textbf{7.88} & \textbf{32.00} & 5.26  & \textbf{6.14} \\
    \midrule
    \textsc{Qwen3-4B} & -     & -     & 87.77 & 82.16 & 32.12 & 22.00 & 27.19 & 9.65 \\
    \;ToolRL & \ding{55}     & \checkmark     & 79.44 & 71.65 & 13.00 & 2.00  & 11.40 & 0.00 \\
    \;Tool-N1 & \ding{55}     & \checkmark     & 88.44 & 77.28 & 34.12 & 24.00 & 9.65  & 11.40 \\
    \;AWM   & \checkmark     & \ding{55}     & 87.12 & 80.61 & 30.00 & 20.00 & 22.81 & 9.65 \\
    \;EnvScaler & \checkmark     & \ding{55}     & 86.50 & \textbf{82.24} & \textbf{38.75} & \textbf{28.00} & \textbf{29.82} & 10.53 \\
    \rowcolor[rgb]{ .851,  .851,  .851} \;TRUSTEE & \checkmark     & \checkmark     & \textbf{88.96} & 81.35 & 35.62 & 22.00 & 28.07 & \textbf{16.67} \\
    \midrule
    \textsc{Qwen3-8B} & -     & -     & 87.77 & 80.31 & 39.88 & 24.00 & 42.98 & 15.79 \\
    \;ToolRL & \ding{55}     & \checkmark     & 88.77 & 81.20 & 41.25 & 22.00 & 16.67 & 14.04 \\
    \;Tool-N1 & \ding{55}     & \checkmark     & 87.83 & 78.90 & 44.38 & 28.00 & 36.84 & 10.53 \\
    \;AWM   & \checkmark     & \ding{55}     & 87.52 & 80.46 & 39.75 & 30.00 & 36.84 & 12.28 \\
    \;EnvScaler & \checkmark     & \ding{55}     & 87.42 & 81.20 & 44.50 & 24.00 & 42.11  & 8.77 \\
    \;Simia-RL & \ding{55}    & \ding{55}     & 32.40 & 46.20 & 2.00  & \textbf{38.00} & 35.09 & 3.51 \\
    \rowcolor[rgb]{ .851,  .851,  .851} \;TRUSTEE & \checkmark     & \checkmark     & \textbf{89.65} & \textbf{82.16} & \textbf{46.38} & 26.00 & \textbf{45.61} & \textbf{17.54} \\
    \bottomrule
    \end{tabular}%
  \label{tab:main}%
\end{table}%

%% file: tables/env_reliability.tex
\begin{wrapfigure}{r}{0.3\textwidth}
\small
  \centering
  \captionof{table}{Simulation quality.}
    \begin{tabular}{lr}
    \toprule
    \textbf{Aspect} & \textbf{Score} \\
    \midrule
    Task Generation & 4.54 \\
    Tool Simulation & 3.77 \\
    User Simulation & 4.64 \\
    Final Reward & 3.90 \\
    Internal Consistency & 4.37 \\
    Overall Fidelity & 3.88 \\
    \bottomrule
    \end{tabular}%
  \label{tab:quality}%
\end{wrapfigure}

%% file: tables/ablation_env.tex
\begin{wrapfigure}{r}{0.55\textwidth}
\centering
\setlength{\tabcolsep}{1pt}
\small
  \captionof{table}{Ablation results on BFCL (\textsc{Qwen3-8B}).}
    \begin{tabular}{lccc}
    \toprule
    \textbf{Setting} & \textbf{Non-Live} & \textbf{Live} & \textbf{Multi-Turn} \\
    \midrule
    \textsc{Qwen3-8B} & 87.77 & 80.31 & 39.88 \\
    TRUSTEE default & \textbf{89.65} & \textbf{82.16} & \textbf{46.38} \\
    \midrule
    \multicolumn{4}{c}{\textit{(I) Environment’s Information Set}} \\
    \midrule
    w/o intent & 88.42 & 80.90 & 43.00 \\
    w/o persona & 86.73 & 79.20 & 43.12 \\
    w/o expected calls & 88.42 & 79.94 & 44.38 \\
    \midrule\midrule
    \multicolumn{4}{c}{\textit{(II) Curriculum Learning}} \\
    \midrule
    w/o curriculum learning & 88.25 & 79.64 & 41.50 \\
    w/o soft curriculum learning & 88.31 & 80.98 & 44.50 \\
    \midrule
    w/o tool & 87.31 & 78.46 & 27.38 \\
    w/o turn & 87.85 & 81.79 & 41.75 \\
    w/o system prompt & 87.56 & 80.46 & 40.75 \\
    w/o user & 87.54 & 79.57 & 32.75 \\
    w/o criteria & 88.42 & 81.05 & 43.12 \\
    \midrule\midrule
    \multicolumn{4}{c}{\textit{(III) Hyper-parameters}} \\
    \midrule
    Number of tools $\uparrow$ & 87.63 & 80.46 & 41.25 \\
    Number of tools $\downarrow$ & 87.35 & 79.79 & 40.25 \\
    \midrule
    Number of turns $\uparrow$ & 87.87 & 80.53 & 44.25 \\
    Number of turns $\downarrow$ & 87.98 & 80.09 & 35.88 \\
    \midrule
    Step size $\uparrow$ & 88.06 & 79.42 & 38.75 \\
    Step size $\downarrow$ & 89.58 & 81.42 & 43.62 \\
    \bottomrule
    \end{tabular}%
    \label{tab:ablation-env}
\end{wrapfigure}

%% file: tables/cost.tex
\begin{wrapfigure}{r}{0.5\textwidth}
\small
  \centering
  \captionof{table}{Training time (hours) on three LMs. The subscript is the relative time to TRUSTEE.}
    \begin{tabular}{lrrr}
    \toprule
    \textbf{Method} & \multicolumn{1}{r}{\textbf{\textsc{8B-Base}}} & \multicolumn{1}{r}{\textbf{\textsc{4B}}} & \multicolumn{1}{r}{\textbf{\textsc{8B}}} \\
    \midrule
    ToolRL & 110$_{2.6\times}$ & 31$_{1.0\times}$ & 21$_{0.7\times}$ \\
    Tool-N1 & 20$_{0.5\times}$ & 27$_{0.9\times}$ & 20$_{0.7\times}$ \\
    EnvScaler & 150$_{3.5\times}$ & 140$_{4.5\times}$ & 170$_{5.7\times}$ \\
    TRUSTEE & 43$_{1.0\times}$ & 31$_{1.0\times}$ & 30$_{1.0\times}$ \\
    \bottomrule
    \end{tabular}%
  \label{tab:cost}%
\end{wrapfigure}

%% file: algorithms/trustee.tex
\begin{algorithm}[htbp]
\footnotesize
\setlength{\abovedisplayskip}{2pt}%
\setlength{\belowdisplayskip}{2pt}%
\caption{\textbf{TRUSTEE}}
\label{alg:trustee}
\begin{algorithmic}[1]
\Require Agent policy $\pi_\theta$; environment backbone LM $\Phi$; tool repository $\mathcal{F}$; number of trajectories per task $M$; batch size $N$; training iterations $T$; initial difficulty $\mathcal{D}_1$; step size $\delta$; reward thresholds $\eta_{\text{high}},\eta_{\text{low}}$; maximum interaction turns $T_{\text{turn}}$; maximum sequence length $L_{\text{seq}}$
\Ensure Updated policy $\pi_\theta$

\For{$n=1$ \textbf{to} $T$}
\Statex \textbf{Stage 1: Task Generation}
\State $\mathcal{B}\gets\emptyset$ \Comment{task batch}
\For{$i=1$ \textbf{to} $N$}
    \State $f_i\gets\zeta(\mathcal{F},\mathcal{D}_n)$;\;$\Omega_G\gets\Omega(f_i,\mathcal{D}_n)$ \Comment{sample tools; build task prompt}
    \State $(p_i,g_i,q_i,c_i)\gets e_0\sim\mathcal{P}_\Phi(\cdot\mid\Omega_G)$ \Comment{generate task}
    \State $h_i\gets\textsc{SystemPrompt}(\mathcal{D}_n)$ \Comment{system prompt per difficulty}
    \State $\mathcal{B}\gets\mathcal{B}\cup\{s_0^{(i)}=(f_i,p_i,g_i,q_i,c_i,h_i,\mathcal{D}_n)\}$ \Comment{initial state}
\EndFor

\Statex \textbf{Stage 2: Simulated Interaction}
\State $\mathcal{T}\gets\emptyset$;\;$\mathcal{R}\gets\emptyset$ \Comment{trajectories; final rewards}
\For{each $s_0^{(i)}\in\mathcal{B}$}
    \For{$m=1$ \textbf{to} $M$}
        \State $\tau\gets(h_i,f_i,q_i)$;\;$t\gets0$ \Comment{initial observation $o_0$}
        \Repeat
            \State $a_t\sim\pi_\theta(\cdot\mid\tau)$ \Comment{agent action}
            \State $\tau\gets\tau\oplus a_t$ \Comment{append action}
            \If{$a_t\in\mathcal{A}_{\text{tool}}$}
                \State $e_t\sim\mathcal{P}_\Phi(\cdot\mid\tau,\Omega_T)$, $\Omega_T=\Omega(f_i,g_i)$ \Comment{tool simulator}
            \Else
                \State $e_t\sim\mathcal{P}_\Phi(\cdot\mid\tau,\Omega_U)$, $\Omega_U=\Omega(f_i,g_i,p_i)$ \Comment{user simulator}
            \EndIf
            \State $\tau\gets\tau\oplus e_t$;\;$t\gets t+1$ \Comment{append feedback}
        \Until{$e_t=\varnothing$ \textbf{or} $t\ge T_{\text{turn}}$ \textbf{or} $|\tau|\ge L_{\text{seq}}$} \Comment{terminate}
        \State $r_{i,m}\sim\mathcal{R}_\Phi(\cdot\mid\tau,\Omega_V)$, $\Omega_V=\Omega(f_i,g_i,p_i,c_i,\mathcal{D}_n)$ \Comment{verify trajectory}
        \State $\mathcal{T}\gets\mathcal{T}\cup\{\tau\}$;\;$\mathcal{R}\gets\mathcal{R}\cup\{r_{i,m}\}$ \Comment{collect}
    \EndFor
\EndFor

\Statex \textbf{Stage 3: Policy Optimization}
\State $\hat{A}\gets\textsc{GroupRelativeAdvantage}(\mathcal{R})$ \Comment{advantage from final reward}
\State $\pi_\theta\gets\textsc{GRPOUpdate}(\pi_\theta,\mathcal{T},\hat{A})$ \Comment{policy optimization}

\Statex \textbf{Difficulty Evolution}
\State $\bar{r}_n\gets\frac{1}{|\mathcal{R}|}\sum_{r\in\mathcal{R}}r$ \Comment{average reward of the batch}
\If{$\bar{r}_n>\eta_{\text{high}}$}
    \State $\mathcal{D}_{n+1}\gets\mathcal{D}_n+\delta$ \Comment{too easy}
\ElsIf{$\bar{r}_n<\eta_{\text{low}}$}
    \State $\mathcal{D}_{n+1}\gets\mathcal{D}_n-\delta$ \Comment{too hard}
\Else
    \State $\mathcal{D}_{n+1}\gets\mathcal{D}_n$ \Comment{unchanged}
\EndIf
\EndFor

\State \Return $\pi_\theta$
\end{algorithmic}
\end{algorithm}

%% file: figures/prompt-task-generation.tex
\begin{figure}
\begin{promptbox}{Prompt Template for Task Generation}
You are an expert at creating realistic tasks for agent training. Given a set of tools and difficulty settings, generate ONE task.\\

\#\# Available Tools

\{tools\}\\

\#\# Settings

- Expected number of distinct tool calls the agent should make: approximately \{number of expected calls\}

- User persona (user's expertise; describe this in the task): \{persona\}

- Query ambiguity (for the first user message only): \{ambiguity\}. User persona and query ambiguity are independent (e.g. an expert can ask vaguely; a novice can ask clearly).

- Expected interaction turns: \{number of turns\}\\

\#\# Your Task

Produce a single task with:

1. Expected turns: The task as a whole should match expected turns: \{number of turns\} Design the goal and first message so the conversation length fits this.

2. expected tool calls: A list of tool names (from the available tools above) that the agent is expected to use, in a plausible order. Length approximately \{number of expected calls\}.

3. user intent: A clear description of what the user ultimately wants to achieve (the ground truth goal for evaluation). This may be richer or more specific than what the user says in the first message.

4. user persona: A short description of the user's expertise level, matching: \{persona\}. Do NOT tie this to query clarity.

5. first user query: The user's first message only—how a real user would actually open the conversation. Keep it natural:

   \quad- It need not state the full intent directly; the user might hint, ask a partial question, or express a vague need that the agent must clarify.
   
   \quad- Match the query ambiguity level: \{ambiguity\}
   
   \quad- Do not name tools; the message should lead naturally toward using the expected tools as the conversation unfolds.
   
   \quad- Avoid sounding like a spec or a list of requirements; write as a real person would (e.g. "I'm trying to figure out..." or "Can you help with something?" rather than "I want you to do X, Y, and Z").\\

\#\# Output Format (JSON only)

\{

    \quad"expected\_tool\_calls": ["tool\_name\_1", "tool\_name\_2", ...],
    
    \quad"user\_intent": "Description of what the user wants to achieve.",
    
    \quad"user\_persona": "Short description of user expertise (not query clarity).",
    
    \quad"first\_user\_query": "The user's first message (ambiguity as specified above)."

\}\\

Generate the task now. Output only valid JSON.
\end{promptbox}
\caption{Prompt template for task generation.}
\label{fig:prompt-task-generation}
\end{figure}

%% file: figures/prompt-tool-simulation.tex
\begin{figure}
\begin{promptbox}{Prompt Template for Tool Simulation}
You are an environment simulator for tool calling agents. Your task is to simulate realistic tool execution results based on the interaction history and the agent's tool call.\\

\#\# Interaction History

\{history\}\\

\#\# Agent's Tool Call

\{tool calling\}\\

\#\# Available Tools

\{tools\}\\

\#\# User Intent (task context)

\{user intent\}\\

\#\# Your Task

Simulate the execution of the agent's tool call and return:

1. Tool Result: The simulated execution output (or an error message if the call is invalid)

2. Tool Reward: An integer reward in \{-1, 0, 1\} evaluating how well this tool call serves the user's intent\\

\#\# Tool Result Guidelines

\#\#\# Validation

First, validate the tool call:

- Parseable: Is the tool call properly formatted and parseable? If the tool call is null or malformed, return an error like "Error: Tool call must be properly formatted in the expected structure"

- Tool exists: Is the tool name in the available tools?

- Required arguments: Are all required parameters provided?

- Argument types: Do argument types match specifications?

- Argument values: Are values realistic and contextually appropriate?

If invalid, return an error message explaining the issue.

\#\#\# Simulation (for valid calls)

If valid, simulate realistic execution output:

- Contextual: Use information from history to generate relevant results

- Realistic: Output should mirror real tool behavior with concrete, specific data

- Consistent: Maintain consistency with previous interactions

- Plausible: Avoid placeholders or generic values; use realistic naming

Examples of good simulation:

- Instead of "file1.txt", use "ProjectReport 2024 Q1.pdf"

- Instead of "user@example.com", use "sarah.chen@techcorp.com"

- Instead of "ID: 123", use "ID: a7c3f91b-2e8d"

Examples of error messages:

- "Error: Tool call is malformed or could not be parsed"

- "Error: Tool 'send email' not found in available tools"

- "Error: Missing required parameter 'recipient' for tool 'send email'"

- "Error: Parameter 'amount' must be a number, got string"\\

\#\# Tool Reward Guidelines

Assign exactly one of -1, 0, or 1:

1 — Tool call fully serves the user's intent

\quad- Valid and well-formed (correct tool, required arguments present and correctly typed)

\quad- Clearly advances the user's goal; optimal or near-optimal choice at this step

\quad- Arguments are accurate and contextually appropriate

0 — Tool call is valid but suboptimal

\quad- Correctly formatted and executable (tool exists, arguments valid)

\quad- Only partly aligned with the user's intent, or a reasonable but not best choice

\quad- May be redundant, slightly off-purpose, or use weaker arguments than possible

-1 — Invalid or intent-mismatched

\quad- Invalid: malformed, unparseable, wrong tool, missing or wrong-typed arguments, or will fail

\quad- Or valid but not addressing the user's intent: wrong task, irrelevant tool, or counterproductive for the goal

First decide validity and intent alignment, then assign the single value \{-1, 0, 1\}. Malformed or unparseable tool calls are always -1.\\

\#\# Output Format

Return your response in the following JSON format:

\{

    \quad"result": "The simulated tool execution output or error message",
    
    \quad"reward": 0
    
\}\\

Important:

- \texttt{result} must be a string (success output or error message; may be JSON inside the string if needed)

- \texttt{reward} must be exactly one of: -1, 0, 1

- Output valid JSON only (no extra text or comments)\\

Generate the simulation now.
\end{promptbox}
    \caption{Prompt template for tool simulation.}
    \label{fig:prompt-tool-simulation}
\end{figure}

%% file: figures/prompt-user-simulation.tex
\begin{figure}
\begin{promptbox}{Prompt Template for User Simulation}
You are simulating a human user interacting with an assistant agent to accomplish a task.\\

\#\# Original Task / First User Message

\{user query\}\\

\#\# User Intent (what the user ultimately wants)

\{user intent\}\\

\#\# User Persona

\{user persona\}\\

\#\# Available Tools (for agent use)

\{tools\}\\

\#\# Interaction History

\{history\}\\

\#\# Agent's Latest Message

\{agent message\}\\

\#\# Your Task

As the user, respond to the agent's latest message and evaluate its quality.

\#\#\# Response Guidelines

Be Human-Like

- Use natural, conversational language

- Be concise and to the point (typically 1-3 sentences)

- Show realistic reactions and emotions when appropriate

- Don't repeat the exact task; use your own words

Provide Information Appropriately

- Only provide information when the agent asks for it

- Don't volunteer all details at once

Know When to End

- If the task is successfully completed, respond with an empty string ""

- If the agent has clearly failed and cannot help, respond with an empty string ""

- Only end when it's genuinely appropriate to stop the conversation

Be Realistic

- React naturally to agent's actions (confusion, satisfaction, frustration, anger, etc.)

- Ask for clarification if the agent's response is unclear

- Point out errors or issues if the agent makes mistakes

\#\#\# Reward Guidelines

Assign exactly one of -1, 0, or 1 based on how well the agent's message serves the user's intent and moves the task forward:

1 — Significant progress; fully meets the user's intent

\quad- Clearly addresses what the user asked or needed at this step

\quad- Moves the task meaningfully toward completion (e.g. correct information, right tool use, useful next step)

\quad- Response is on-topic, clear, and appropriate

0 — Little progress or suboptimal

\quad- Partly relevant or only marginally helpful

\quad- Some progress but slow, indirect, or with unnecessary detours

\quad- Suboptimal choices (e.g. weaker tool use or incomplete answer) but not wrong

-1 — Wrong, unclear, no progress, or off-topic

\quad- Wrong answer, off-topic, or fails to address the user's intent

\quad- Unclear or confusing so the user cannot act on it

\quad- No real progress (e.g. repetition, irrelevant actions, or counterproductive steps)

First judge progress and intent alignment, then assign the single value \{-1, 0, 1\}.\\

\#\# Output Format

Return your response in the following JSON format:

\{

    "response": "Your message as the user",
    
    "reward": 0
    
\}\\

Important:

- \texttt{response} must be a string; use an empty string "" if the conversation should end

- \texttt{reward} must be exactly one of: -1, 0, 1

- Output valid JSON only (no extra text or comments)\\

Generate the user simulation now.
\end{promptbox}
\caption{Prompt template for user simulation.}
\label{fig:prompt-user-simulation}
\end{figure}

%% file: figures/prompt-verifer-simulation.tex
\begin{figure}
\begin{promptbox}{Prompt Template for Verifier Simulation}
Your task is to evaluate how well an assistant agent performed in helping a user accomplish their task.\\

\#\# User's Original Query

\{user query\}\\

\#\# Expected Tool Calls (should be used correctly)

\{expected tool calls\}\\

\#\# User Intent (ground truth)

\{user intent\}\\

\#\# User Persona

\{user persona\}\\

\#\# Available Tools

\{tools text\}\\

\#\# Complete Interaction Trajectory

\{interaction history text\}\\

\#\# Evaluation Task

Evaluate the assistant agent's overall performance using ONLY the following \{num criteria\} key criteria (in order of importance):

\{criteria text\}

\#\#\# Reward Scale

Assign a final reward of 1, 0, or -1:

- 1: All of the above \{num criteria\} key criteria are met.

- 0: Most of the above \{num criteria\} key criteria are met.

- -1: Most of the above \{num criteria\} key criteria are NOT met or the agent does not call the tools at all.\\

\#\# Output Format

Return your evaluation in the following JSON format:

\{

    "reward": 0,
    
    "reasoning": "Brief explanation (1-2 sentences)"
    
\}\\

Important:

- \texttt{reward} must be exactly one of: -1, 0, 1

- Output valid JSON only\\

Generate the evaluation now.
\end{promptbox}
\caption{Prompt template for verifier simulation.}
\label{fig:prompt-verifier-simulation}
\end{figure}

%% file: tables/difficulty.tex
\begin{table}[htbp]
\small
  \centering
  \caption{Details of task difficulty.}
    \begin{tabularx}{\linewidth}{ccX}
    \toprule
    \textbf{Aspect} & \textbf{Level} & \textbf{Description} \\
    \midrule
    \multirow{3}[12]{*}{Interaction Turns} & 0     & One or a couple of turns: the task should be resolvable in a short exchange (e.g. 1–2 back-and-forths). \\
\cmidrule{2-3}          & 1     & A few to several turns: the task should naturally involve some back-and-forth (e.g. clarification, multiple steps, a handful of exchanges). \\
\cmidrule{2-3}          & 2     & Many turns: the task should require an extended conversation (e.g. multi-step workflow, several clarifications, or many tool uses and replies). \\
    \midrule
    \multirow{4}[8]{*}{System Prompt} & 0     & See Figure~\ref{fig:system-prompt-detailed}. \\
\cmidrule{2-3}          & 1     & See Figure~\ref{fig:system-prompt-concise}. \\
\cmidrule{2-3}          & 2     & You are a helpful multi-turn dialogue assistant capable of leveraging tool calls to solve user tasks. \\
\cmidrule{2-3}          & 3     & You are a helpful assistant. \\
    \midrule\midrule
    \multirow{3}[4]{*}{User Persona} & 0     & Expert: the user has extensive prior knowledge in the domain. \\
\cmidrule{2-3}          & 1     & Beginner: the user has some prior knowledge but may need guidance. \\
\cmidrule{2-3}          & 2     & Novice: the user has no prior knowledge and may need clarification on basics. \\
    \midrule
    \multirow{3}[9]{*}{Query Ambiguity} & 0     & Clear: the first user message should be specific, complete, and unambiguous. \\
\cmidrule{2-3}          & 1     & Somewhat ambiguous: the first user message may omit some details or be a little vague. \\
\cmidrule{2-3}          & 2     & Highly ambiguous: the first user message should be vague, incomplete, or missing key information that the agent may need to clarify. \\
    \midrule
    \multirow{5}[12]{*}{Rewarding Criteria} & 0     & Meet the user's intent: Did the agent fulfill what the user wanted by using the tools? \\
\cmidrule{2-3}          & 1     & Correctly use the expected tools: Did the agent use the expected tools appropriately and correctly? \\
\cmidrule{2-3}          & 2     & Free of hallucination: No invented facts, tool names, or outputs. \\
\cmidrule{2-3}          & 3     & No redundant or incorrect trials: No unnecessary or wrong tool calls or steps. \\
\cmidrule{2-3}          & 4     & Concise and efficient: Response and tool use are to-the-point and efficient. \\
    \bottomrule
    \end{tabularx}%
  \label{tab:difficulty}%
\end{table}%

%% file: figures/system-prmopt-detailed.tex
\begin{figure}
\begin{promptbox}{Detailed System Prompt}
You are a helpful multi-turn dialogue assistant capable of leveraging tool calls to solve user tasks.\\

\# Steps for Each Turn

1. Think: Recall relevant context and analyze the current user goal.

2. Decide on Tool Usage: If a tool is needed, specify the tool and its parameters.

3. Respond Appropriately: If a response is needed, generate one while maintaining consistency across user queries.\\

\# Output Format

If you need to call tools, use the following format:

\quad\texttt{<think> Your thoughts and reasoning </think>}

\quad\texttt{<tool\_call> Tool call JSON object(s) </tool\_call>}

If you need to respond to the user, use the following format:

\quad\texttt{<think> Your thoughts and reasoning </think>}

\quad\texttt{Your response to the user.}\\

\# Important Notes

1. You must always include the \texttt{<think>} field to outline your reasoning. Provide either calling tools or responding to the user. The former leads to an interaction with the tool environment, while the latter switches back to the user interface.

2. You may invoke multiple tool calls simultaneously in the \texttt{<tool\_call>} field. You will get the tool feedback after calling the tools. You may call tools for multiple turns if needed.

3. Refer to the previous dialogue records in the history, including the user's queries, previous tool calls, tool feedback, and your responses.
\end{promptbox}
\caption{Detailed system prompt.}
\label{fig:system-prompt-detailed}
\end{figure}

%% file: figures/system-prompt-concise.tex
\begin{figure}
\begin{promptbox}{Concise System Prompt}
You are a helpful multi-turn dialogue assistant capable of leveraging tool calls to solve user tasks.\\

\# Important Notes

1. You must always include the \texttt{<think>} field to outline your reasoning. Provide either calling tools or responding to the user. The former leads to an interaction with the tool environment, while the latter switches back to the user interface.

2. You may invoke multiple tool calls simultaneously in the \texttt{<tool\_call>} field. You will get the tool feedback after calling the tools. You may call tools for multiple turns if needed.

3. Refer to the previous dialogue records in the history, including the user's queries, previous tool calls, tool feedback, and your responses.
\end{promptbox}
\caption{Concise system prompt.}
\label{fig:system-prompt-concise}
\end{figure}

%% file: figures/prompt-tool-filter.tex
\begin{figure}
\begin{promptbox}{Prompt Template for Tool Preprocessing}
You are a tool quality evaluator. Determine if this tool should be kept based on:

\#\# Evaluation Criteria

1. Tool Name: Must be meaningful (not generic like "api", "hello", "test")
2. Description: Must be informative (not just repeating the name or single letters)
3. Parameters: Must have reasonable descriptions (not just "Parameter: \{name\}")
4. Overall Quality: Tool should be useful and clear

\#\# Tool to Evaluate

\{tool\}

\#\# Response Format

Respond ONLY with "KEEP" or "DISCARD" followed by a brief reason.

Format: KEEP/DISCARD: reason
\end{promptbox}
\caption{Prompt template for tool preprocessing.}
\label{fig:prompt-tool-filter}
\end{figure}

%% file: tables/hyper-parameters.tex
\begin{table}[htbp]
\small
  \centering
  \caption{Hyper-parameters for training.}
    \begin{tabular}{lc}
    \toprule
    \textbf{Hyper-parameter} & \textbf{Value} \\
    \midrule
    Learning Rate & 1e-6 \\
    Total Steps & 200 \\
    Batch Size (Number of Tasks per Step) & 16 \\
    Mini Batch Size & 4 \\
    Micro Batch Size & 1 \\
    KL Loss Coefficient & 0.001 \\
    Clip Ratio & 0.2 \\
    Temperature & 1.0 \\
    Group Size & 8 \\
    Maximum Prompt Length & 8,192 \\
    Maximum Response Length & 16,384 \\
    \bottomrule
    \end{tabular}%
  \label{tab:hyper-parameters}%
\end{table}%

%% file: tables/setting_hparams.tex
\begin{table}[htbp]
\small
  \centering
  \caption{Setting for different levels of hyper-parameters.}
    \begin{tabular}{lccccc}
    \toprule
    \textbf{Setting} & \textbf{Tool} & \textbf{User Turn} & \textbf{Tool Turn} & \textbf{Agent Turn} & \textbf{$\delta$} \\
    \midrule
    Default & 10    & 3     & 3     & 6     & 3 \\
    \midrule
    Tool$\uparrow$ & 20    & 3     & 3     & 6     & 3 \\
    Tool$\downarrow$ & 5     & 3     & 3     & 6     & 3 \\
    \midrule
    Turn$\uparrow$ & 10    & 5     & 5     & 10    & 3 \\
    Turn$\downarrow$ & 10    & 1     & 1     & 2     & 3 \\
    \midrule
    Step$\uparrow$ & 10    & 3     & 3     & 6     & 10 \\
    Step$\downarrow$ & 10    & 3     & 3     & 6     & 1 \\
    \bottomrule
    \end{tabular}%
  \label{tab:setting-hparam}%
\end{table}%

%% file: tables/ci.tex
\begin{table}[htbp]
\tiny
\setlength{\tabcolsep}{3pt}
  \centering
  \caption{Main results with confidence intervals.}
    \begin{tabular}{lcccccc}
    \toprule
    \textbf{Method} & \textbf{BFCL (Non-Live)} & \textbf{BFCL (Live)} & \textbf{BFCL (Multi-Turn)} & \textbf{$\tau^2$-bench (Airline)} & \textbf{$\tau^2$-bench (Retail)} & \textbf{$\tau^2$-bench (Telecom)} \\
    \midrule
    \textsc{Qwen3-8B-Base} & 8.38 [6.82, 10.03] & 4.74 [3.66, 6.08] & 0.00 [0.00, 0.48] & 26.00 [15.9, 39.6] & 4.39 [1.9, 9.9] & 1.75 [0.5, 6.2] \\
    ToolRL & 40.83 [38.17, 43.66] & 56.33 [53.67, 58.94] & 0.25 [0.00, 0.97] & 30.00 [19.1, 43.8] & 6.14 [3.0, 12.1] & 5.26 [2.4, 11.0] \\
    Tool-N1 & 37.85 [34.85, 40.81] & 29.53 [27.16, 32.02] & 0.12 [0.02, 0.70] & 32.00 [20.8, 45.8] & 4.39 [1.9, 9.9] & 4.39 [1.9, 9.9] \\
    EnvScaler & 75.29 [72.46, 78.06] & 62.03 [59.41, 64.58] & 0.62 [0.27, 1.45] & 28.00 [17.5, 41.7] & 6.14 [3.0, 12.1] & 0.00 [0.0, 3.3] \\
    TRUSTEE & 83.92 [81.44, 86.25] & 72.54 [70.10, 74.85] & 7.88 [6.20, 9.95] & 32.00 [20.8, 45.8] & 5.26 [2.4, 11.0] & 6.14 [3.0, 12.1] \\
    \midrule
    \textsc{Qwen3-4B} & 87.77 [85.56, 89.85] & 82.16 [80.03, 84.11] & 32.12 [28.98, 35.44] & 22.00 [12.8, 35.2] & 27.19 [19.9, 36.0] & 9.65 [5.5, 16.5] \\
    ToolRL & 79.44 [76.75, 82.08] & 71.65 [69.19, 73.99] & 13.00 [10.85, 15.51] & 2.00 [0.4, 10.5] & 11.40 [6.8, 18.5] & 0.00 [0.0, 3.3] \\
    Tool-N1 & 88.44 [86.38, 90.42] & 77.28 [74.97, 79.43] & 34.12 [30.92, 37.48] & 24.00 [14.3, 37.4] & 9.65 [5.5, 16.5] & 11.40 [6.8, 18.5] \\
    AWM   & 87.12 [84.96, 89.23] & 80.61 [78.41, 82.63] & 30.00 [26.93, 33.26] & 20.00 [11.2, 33.0] & 22.81 [16.1, 31.3] & 9.65 [5.5, 16.5] \\
    EnvScaler & 86.50 [84.23, 88.67] & 82.24 [80.11, 84.18] & 38.75 [35.44, 42.17] & 28.00 [17.5, 41.7] & 29.82 [22.2, 38.8] & 10.53 [6.1, 17.5] \\
    TRUSTEE & 88.96 [86.90, 90.96] & 81.35 [79.18, 83.33] & 35.62 [32.38, 39.00] & 22.00 [12.8, 35.2] & 28.07 [20.6, 36.9] & 16.67 [10.9, 24.6] \\
    \midrule
    \textsc{Qwen3-8B} & 87.77 [85.65, 89.77] & 80.31 [78.11, 82.34] & 39.88 [36.54, 43.31] & 24.00 [14.3, 37.4] & 42.98 [34.3, 52.2] & 15.79 [10.2, 23.6] \\
    ToolRL & 88.77 [86.67, 90.75] & 81.20 [79.03, 83.19] & 41.25 [37.89, 44.70] & 22.00 [12.8, 35.2] & 16.67 [10.9, 24.6] & 14.04 [8.8, 21.6] \\
    Tool-N1 & 87.83 [85.71, 89.88] & 78.90 [76.65, 81.00] & 44.38 [40.97, 47.84] & 28.00 [17.5, 41.7] & 36.84 [28.6, 46.0] & 10.53 [6.1, 17.5] \\
    AWM   & 87.52 [85.35, 89.62] & 80.46 [78.26, 82.49] & 39.75 [36.42, 43.18] & 30.00 [19.1, 43.8] & 36.84 [28.6, 46.0] & 12.28 [7.5, 19.6] \\
    EnvScaler & 87.42 [85.25, 89.54] & 81.20 [79.03, 83.19] & 44.50 [41.09, 47.96] & 22.00 [12.8, 35.2] & 4.39 [1.9, 9.9] & 9.65 [5.5, 16.5] \\
    TRUSTEE & 89.65 [87.67, 91.56] & 82.16 [80.03, 84.11] & 46.38 [42.94, 49.84] & 26.00 [15.9, 39.6] & 45.61 [36.8, 54.8] & 17.54 [11.7, 25.6] \\
    \bottomrule
    \end{tabular}%
  \label{tab:ci}%
\end{table}%

%% file: tables/adv.tex
\begin{wrapfigure}{r}{0.5\textwidth}
\small
\centering
\setlength{\tabcolsep}{3pt}
\small
  \captionof{table}{Analysis on the advantage.}
    \begin{tabular}{lccc}
    \toprule
    \textbf{Advantage} & \textbf{Non-Live} & \textbf{Live} & \textbf{Multi-Turn} \\
    \midrule
    Sequence-level & \textbf{88.96} & 81.35 & \textbf{35.62} \\
    Turn-level & 88.31 & \textbf{81.94} & 33.00 \\
    \bottomrule
    \end{tabular}%
    \label{tab:adv}
\end{wrapfigure}

%% file: tables/teacher.tex
\begin{table}[htbp]
\small
  \centering
  \caption{Analysis on the environment backbone LM.}
    \begin{tabular}{llccc}
    \toprule
    \textbf{Agent LM $\theta$} & \textbf{Environment LM $\Phi$} & \textbf{Non-Live} & \textbf{Live} & \textbf{Multi-Turn} \\
    \midrule
    \multirow{2}{*}{\textsc{Qwen3-8B-Base}} & \textsc{Qwen3-8B} & \textbf{83.92} & 72.54 & \textbf{7.88} \\
          & \textsc{Qwen3.5-397B-A17B-FP8} & 78.42 & \textbf{73.58} & 2.38 \\
    \midrule
    \multirow{2}{*}{\textsc{Qwen3-4B}} & \textsc{Qwen3-8B} & \textbf{88.96} & \textbf{81.35} & 35.62 \\
          & \textsc{Qwen3.5-397B-A17B-FP8} & 87.23 & 80.01 & \textbf{37.88} \\
    \midrule
    \multirow{2}{*}{\textsc{Qwen3-8B}} & \textsc{Qwen3-8B} & \textbf{89.65} & \textbf{82.16} & \textbf{46.38} \\
          & \textsc{Qwen3.5-397B-A17B-FP8} & 88.35 & 79.20 & 41.25 \\
    \bottomrule
    \end{tabular}%
  \label{tab:teacher}%
\end{table}%

%% file: tables/stateful.tex
\begin{wrapfigure}{r}{0.56\textwidth}
\small
  \centering
  \captionof{table}{Performance of stateful tool simulation on BFCL with \textsc{Qwen3-8B-Base}.}
    \begin{tabular}{lrrr}
    \toprule
    \textbf{Setting} & \textbf{Non-Live} & {\textbf{Live}} & {\textbf{Multi-Turn}} \\
    \midrule
    Qwen3-8B-Base & 8.38  & 4.74  & 0.00 \\
    TRUSTEE (default) & \textbf{83.92} & \textbf{72.54} & \textbf{7.88} \\
    TRUSTEE (stateful) & 74.60 & 51.15 & 1.00 \\
    \bottomrule
    \end{tabular}%
  \label{tab:stateful}%
\end{wrapfigure}

%% file: tables/license.tex
\begin{table}[htbp]
\footnotesize
  \centering
  \caption{URLs and licenses of scientific artifacts we use.}
    \begin{tabular}{lll}
    \toprule
    \textbf{Artifact} & \textbf{URL} & \textbf{License} \\
    \midrule
    Qwen3 Models & \url{https://huggingface.co/collections/Qwen/qwen3} & Apache-2.0 \\
    Qwen3.5 Models & \url{https://huggingface.co/collections/Qwen/qwen35} & Apache-2.0 \\
    verl  & \url{https://github.com/volcengine/verl} & Apache-2.0 \\
    vLLM  & \url{https://github.com/vllm-project/vllm} & Apache-2.0 \\
    BFCL  & \url{https://github.com/ShishirPatil/gorilla} & Apache-2.0 \\
    $\tau^2$-bench & \url{https://github.com/sierra-research/tau2-bench} & MIT \\
    ToolBench & \url{https://github.com/openbmb/toolbench} & Apache-2.0 \\
    ToolRL & \url{https://github.com/qiancheng0/ToolRL} & Apache-2.0 \\
    Tool-N1 & \url{https://github.com/NVlabs/Tool-N1} & Apache-2.0 \\
    EnvScaler & \url{https://github.com/RUC-NLPIR/EnvScaler} & MIT \\
    AWM   & \url{https://huggingface.co/collections/Snowflake/agent-world-model} & Apache-2.0 \\
    Simia-RL   & \url{https://huggingface.co/Simia-Agent/Simia-Tau-SFT-RL-Qwen3-8B} & MIT \\
    \bottomrule
    \end{tabular}%
  \label{tab:license}%
\end{table}%

%% file: checklist.tex
\section*{NeurIPS Paper Checklist}

\begin{enumerate}

\item {\bf Claims}
    \item[] Question: Do the main claims made in the abstract and introduction accurately reflect the paper's contributions and scope?
    \item[] Answer: \answerYes{} 
    \item[] Justification: The contributions and scope of this paper are accurately reflected by the main claims made in the abstract and introduction.

\item {\bf Limitations}
    \item[] Question: Does the paper discuss the limitations of the work performed by the authors?
    \item[] Answer: \answerYes{} 
    \item[] Justification: Discussed in Section~\ref{sec:limitation-impact}.

\item {\bf Theory assumptions and proofs}
    \item[] Question: For each theoretical result, does the paper provide the full set of assumptions and a complete (and correct) proof?
    \item[] Answer: \answerNA{} 
    \item[] Justification: This paper does not include theoretical results (\textit{i.e.}, theorems and proofs), but only provides theoretical analysis with conceptual propositions and justifications (Section~\ref{subsec:theory} and Appendix~\ref{app:theory}).

    \item {\bf Experimental result reproducibility}
    \item[] Question: Does the paper fully disclose all the information needed to reproduce the main experimental results of the paper to the extent that it affects the main claims and/or conclusions of the paper (regardless of whether the code and data are provided or not)?
    \item[] Answer: \answerYes{} 
    \item[] Justification: Details for reproduction are provided in Section~\ref{sec:setup}, Appendix~\ref{app:prompt} and \ref{app:implementation-detail}.

\item {\bf Open access to data and code}
    \item[] Question: Does the paper provide open access to the data and code, with sufficient instructions to faithfully reproduce the main experimental results, as described in supplemental material?
    \item[] Answer: \answerYes{} 
    \item[] Justification: We have provided the URL to our code and data in Section~\ref{sec:intro}.

\item {\bf Experimental setting/details}
    \item[] Question: Does the paper specify all the training and test details (e.g., data splits, hyperparameters, how they were chosen, type of optimizer) necessary to understand the results?
    \item[] Answer: \answerYes{} 
    \item[] Justification: Provided in Section~\ref{sec:setup}, Appendix~\ref{app:prompt} and \ref{app:implementation-detail}.

\item {\bf Experiment statistical significance}
    \item[] Question: Does the paper report error bars suitably and correctly defined or other appropriate information about the statistical significance of the experiments?
    \item[] Answer: \answerNo{} 
    \item[] Justification: It is highly computationally expensive to train and evaluate the models. We cannot afford to run multiple times to report statistical significance. Yet we provide the confidence intervals of our main results in Appendix~\ref{app:ci}, which is not run-to-run.

\item {\bf Experiments compute resources}
    \item[] Question: For each experiment, does the paper provide sufficient information on the computer resources (type of compute workers, memory, time of execution) needed to reproduce the experiments?
    \item[] Answer: \answerYes{} 
    \item[] Justification: Provided in Section~\ref{subsec:cost} and Appendix~\ref{app:compute}.
    
\item {\bf Code of ethics}
    \item[] Question: Does the research conducted in the paper conform, in every respect, with the NeurIPS Code of Ethics \url{https://neurips.cc/public/EthicsGuidelines}?
    \item[] Answer: \answerYes{} 
    \item[] Justification: This paper conforms with the NeurIPS Code of Ethics in every respect.

\item {\bf Broader impacts}
    \item[] Question: Does the paper discuss both potential positive societal impacts and negative societal impacts of the work performed?
    \item[] Answer: \answerYes{} 
    \item[] Justification: Discussed in Section~\ref{sec:limitation-impact}.
    
\item {\bf Safeguards}
    \item[] Question: Does the paper describe safeguards that have been put in place for responsible release of data or models that have a high risk for misuse (e.g., pre-trained language models, image generators, or scraped datasets)?
    \item[] Answer: \answerNA{} 
    \item[] Justification: To the best of our knowledge, this paper poses no major risks.

\item {\bf Licenses for existing assets}
    \item[] Question: Are the creators or original owners of assets (e.g., code, data, models), used in the paper, properly credited and are the license and terms of use explicitly mentioned and properly respected?
    \item[] Answer: \answerYes{} 
    \item[] Justification: Discussed in Appendix~\ref{app:license}.

\item {\bf New assets}
    \item[] Question: Are new assets introduced in the paper well documented and is the documentation provided alongside the assets?
    \item[] Answer: \answerNA{} 
    \item[] Justification: This paper does not release new assets.

\item {\bf Crowdsourcing and research with human subjects}
    \item[] Question: For crowdsourcing experiments and research with human subjects, does the paper include the full text of instructions given to participants and screenshots, if applicable, as well as details about compensation (if any)? 
    \item[] Answer: \answerNA{} 
    \item[] Justification: This paper does not involve crowdsourcing nor research with human subjects.

\item {\bf Institutional review board (IRB) approvals or equivalent for research with human subjects}
    \item[] Question: Does the paper describe potential risks incurred by study participants, whether such risks were disclosed to the subjects, and whether Institutional Review Board (IRB) approvals (or an equivalent approval/review based on the requirements of your country or institution) were obtained?
    \item[] Answer: \answerNA{} 
    \item[] Justification: This paper does not involve crowdsourcing nor research with human subjects.

\item {\bf Declaration of LLM usage}
    \item[] Question: Does the paper describe the usage of LLMs if it is an important, original, or non-standard component of the core methods in this research? Note that if the LLM is used only for writing, editing, or formatting purposes and does \emph{not} impact the core methodology, scientific rigor, or originality of the research, declaration is not required.
    \item[] Answer: \answerYes{} 
    \item[] Justification: Discussed in Appendix~\ref{app:llm}.

\end{enumerate}